\documentclass[pdflatex,sn-basic]{sn-jnl}

\usepackage{graphicx}%
\usepackage{multirow}%
\usepackage{amsmath,amssymb,amsfonts}%
\usepackage{amsthm}%
\usepackage{mathrsfs}%
\usepackage[title]{appendix}%
\usepackage{xcolor}%
\usepackage{textcomp}%
\usepackage{manyfoot}%
\usepackage{booktabs}%
\usepackage{algorithm}%
\usepackage{algorithmicx}%
\usepackage{algpseudocode}%
\usepackage{listings}%
\usepackage[utf8x]{inputenc}
\usepackage[T1]{fontenc}
\usepackage[polish, english]{babel}
\usepackage{diagbox}
\usepackage{hyperref}
\usepackage{tabularx}
\usepackage{bm}
\usepackage{caption}
\usepackage{geometry}
\usepackage{subcaption}
\usepackage{lipsum}
\usepackage[table]{xcolor}
\usepackage{adjustbox}
\usepackage{calc}
\usepackage{tikz}
\usetikzlibrary{spy}

\usepackage{todonotes}
\setuptodonotes{inline}

\theoremstyle{thmstyleone}%
\theoremstyle{thmstyletwo}%

\theoremstyle{thmstylethree}%

\begin{document}

\title{Semi-Supervised Diversity-Aware Domain Adaptation for 3D Object detection}


\author[1]{\fnm{Jakub} \sur{Winter}}
\email{jakub.winter.stud@pw.edu.pl}

\author[2]{\fnm{Paweł} \sur{Wawrzyński}}
\email{pawel.wawrzynski@ideas-ncbr.pl}

\author[1]{\fnm{Andrii} \sur{Gamalii}}
\email{andrii.gamalii.stud@pw.edu.pl}

\author[1]{\fnm{Daniel} \sur{Górniak}}
\email{daniel.gorniak@pw.edu.pl}

\author[1]{\fnm{Bartłomiej} \sur{Olber}}
\email{bartlomiej.olber@pw.edu.pl}

\author[1]{\fnm{Marcin} \sur{Łojek}}
\email{marcin.lojek.stud@pw.edu.pl}

\author[1]{\fnm{Robert Marek} \sur{Nowak}}
\email{robert.nowak@pw.edu.pl}

\author*[1]{\fnm{Krystian} \sur{Radlak}}
\email{krystian.radlak@pw.edu.pl}

\affil*[1]{\orgname{Warsaw University of Technology},
           \country{Poland}}

\affil[2]{\orgname{IDEAS NCBR},
           \country{Poland}}

\abstract{3D object detectors are fundamental components of perception systems in autonomous vehicles. While these detectors achieve remarkable performance on standard autonomous driving benchmarks, they often struggle to generalize across different domains—for instance, a model trained in the U.S. may perform poorly in regions like Asia or Europe. This paper presents a novel LiDAR domain adaptation method based on neuron activation patterns, demonstrating that state-of-the-art performance can be achieved by annotating only a small, representative, and diverse subset of samples from the target domain if they are correctly selected. The proposed approach requires very small annotation budget and, when combined with post-training techniques  inspired by continual learning prevent weight drift from the original model. Empirical evaluation shows that the proposed domain adaptation approach outperforms both linear probing and state-of-the-art domain adaptation techniques.}


\maketitle

\def\wei{{\bm w}}

\section{Introduction}
Recent advancements in the fields of computer vision and artificial intelligence have sparked the interest of researchers and the general public in autonomous driving. This curiosity has led to the development of numerous publicly available open datasets. The most popular and widely used of these datasets are KITTI \cite{kitti_dataset}, NuScenes \cite{nuscenes_dataset} and Waymo \cite{waymo_dataset}. Lesser known or newer datasets include Lyft \cite{lyft_dataset}, Argoverse \cite{argoverse_dataset}, Zenseact \cite{zenseact_dataset}, Pandaset \cite{xiao2021pandaset} and ONCE \cite{mao2021one}. These datasets include annotated road traffic data recorded by  LiDARs and cameras. Availability of such data has enabled researchers to develop increasingly better object detection methods.

\newcommand{\referencecamimage}{\includegraphics[height=2.5cm]{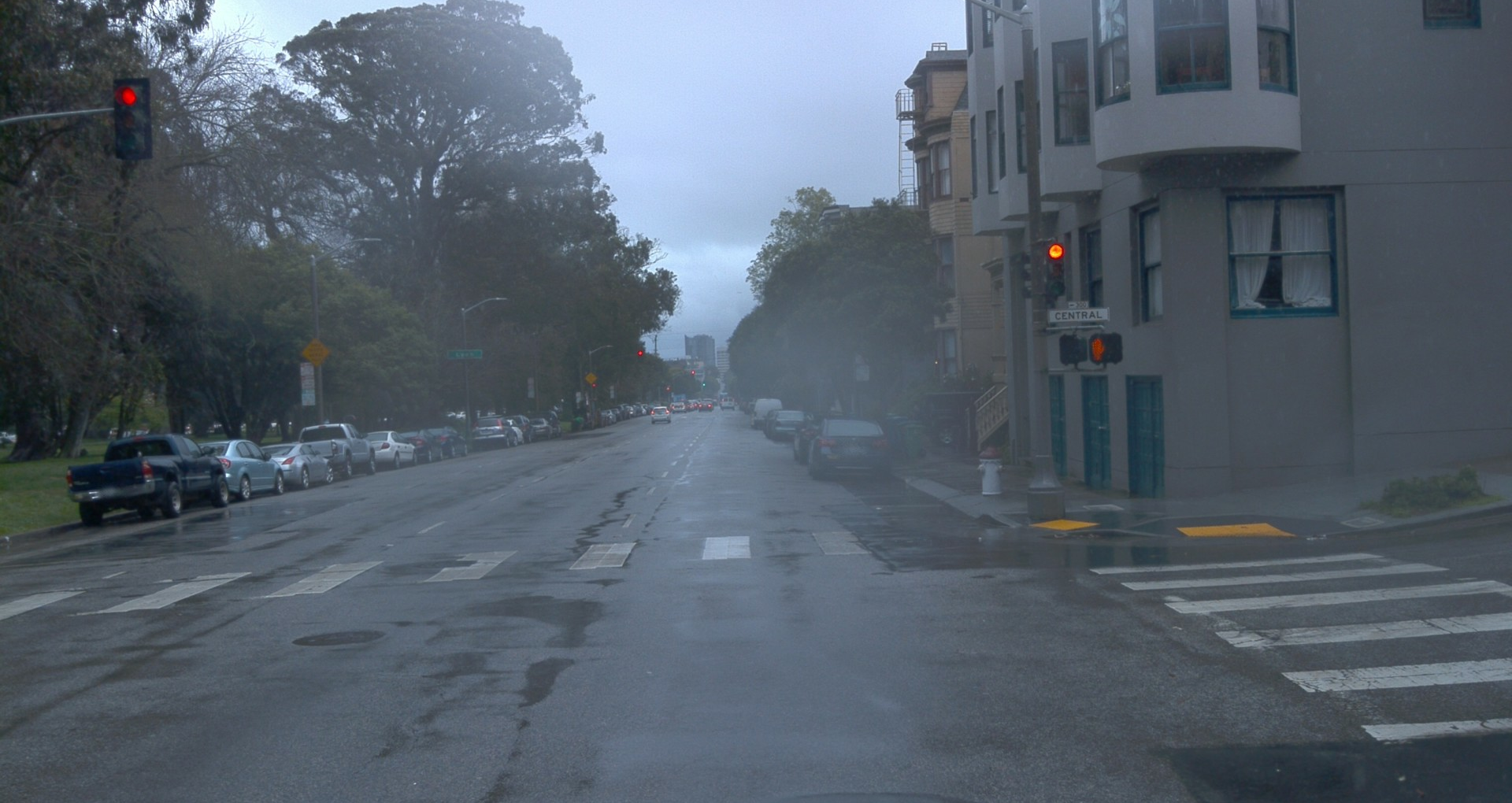}}
\newcommand{\referencepcimage}{\includegraphics[height=2.5cm]{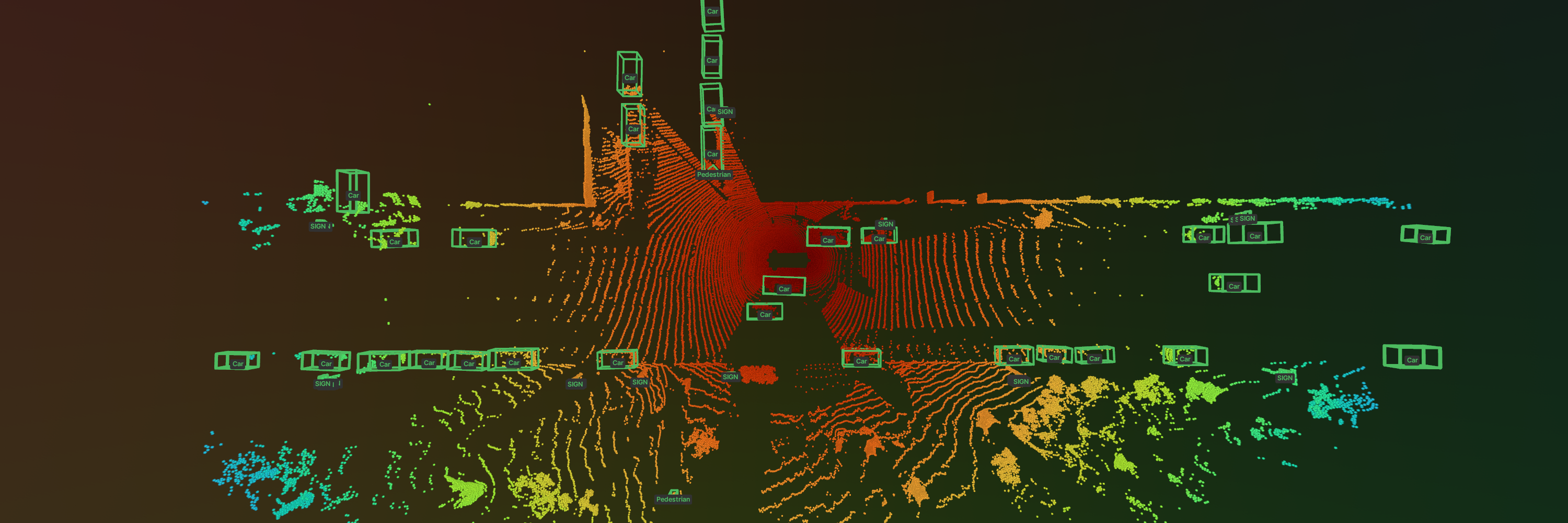}}

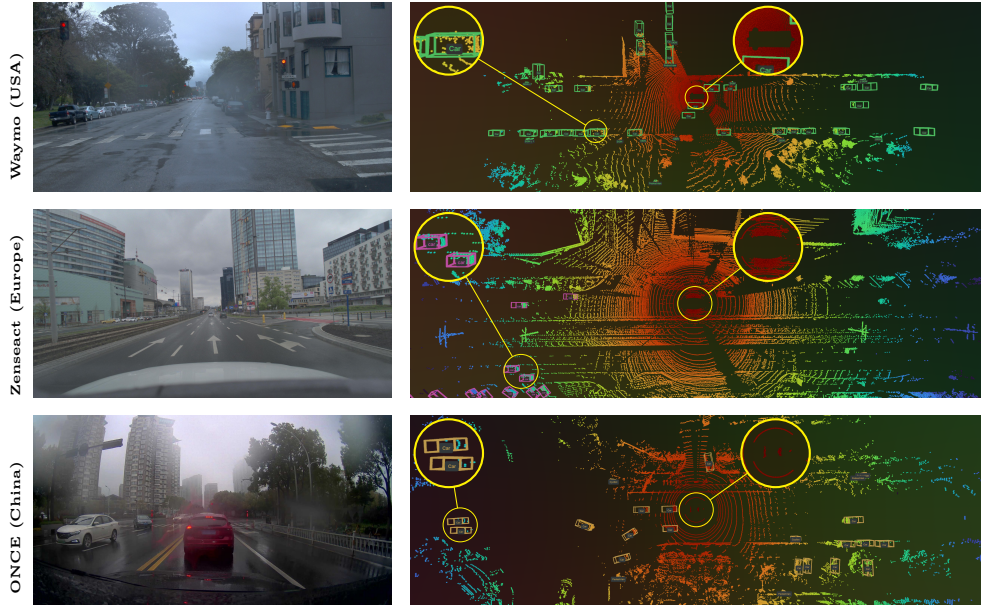
\begin{figure}[t!]
\centering
\tabcolsep=0.01cm
\renewcommand{\arraystretch}{-1.5}
    \begin{tabular}{@{}ccc@{}}
    \centering
        \rotatebox{90}{\textbf{\hspace{0.13cm} \tiny Waymo (USA)}} 
        &
        \begin{tikzpicture}[spy using outlines={circle,yellow,magnification=3,size=0.9cm, connect spies}]
            \node {\referencecamimage};
        \end{tikzpicture}
        & 
        \begin{tikzpicture}[spy using outlines={circle,yellow,magnification=3,size=0.9cm, connect spies}]
            \node {\referencepcimage};
            \spy on (0.01,0) in node [right] at (0.5,0.75);
            \spy on (-1.32,-0.45) in node [right] at (-3.7,0.75);
        \end{tikzpicture}
        \\

        \rotatebox{90}{\textbf{\hspace{0.001cm} \tiny Zenseact (Europe)}} 
        &
        \begin{tikzpicture}[spy using outlines={circle,yellow,magnification=3,size=0.9cm, connect spies}]
            \node {\includegraphics[height=\heightof{\referencecamimage}, width=\widthof{\referencecamimage}]
                    {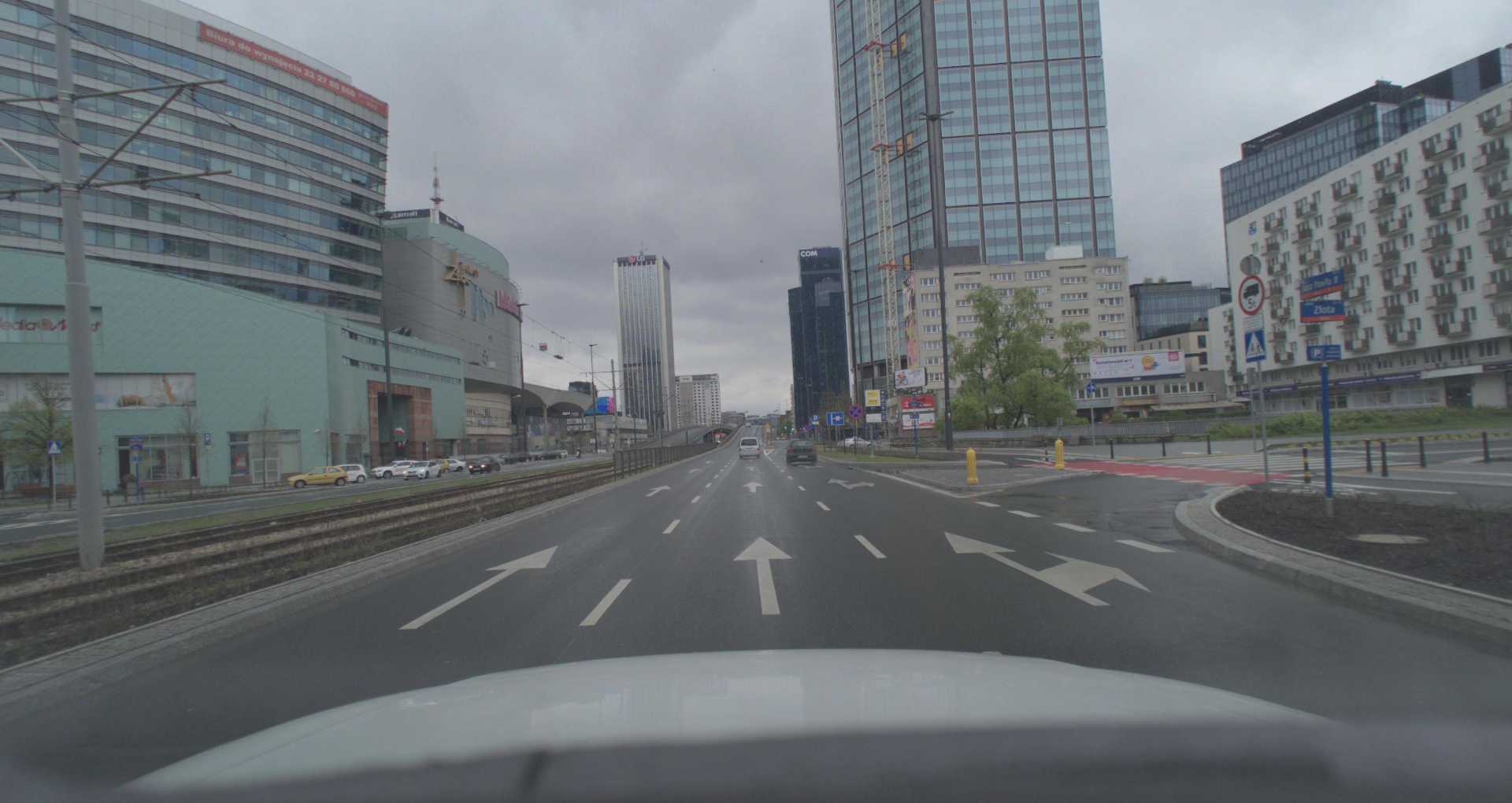}};
        \end{tikzpicture}
        & 
        \begin{tikzpicture}[spy using outlines={circle,yellow,magnification=2,size=0.9cm, connect spies}]
            \node {\includegraphics[height=\heightof{\referencepcimage}, width=\widthof{\referencepcimage}]
                    {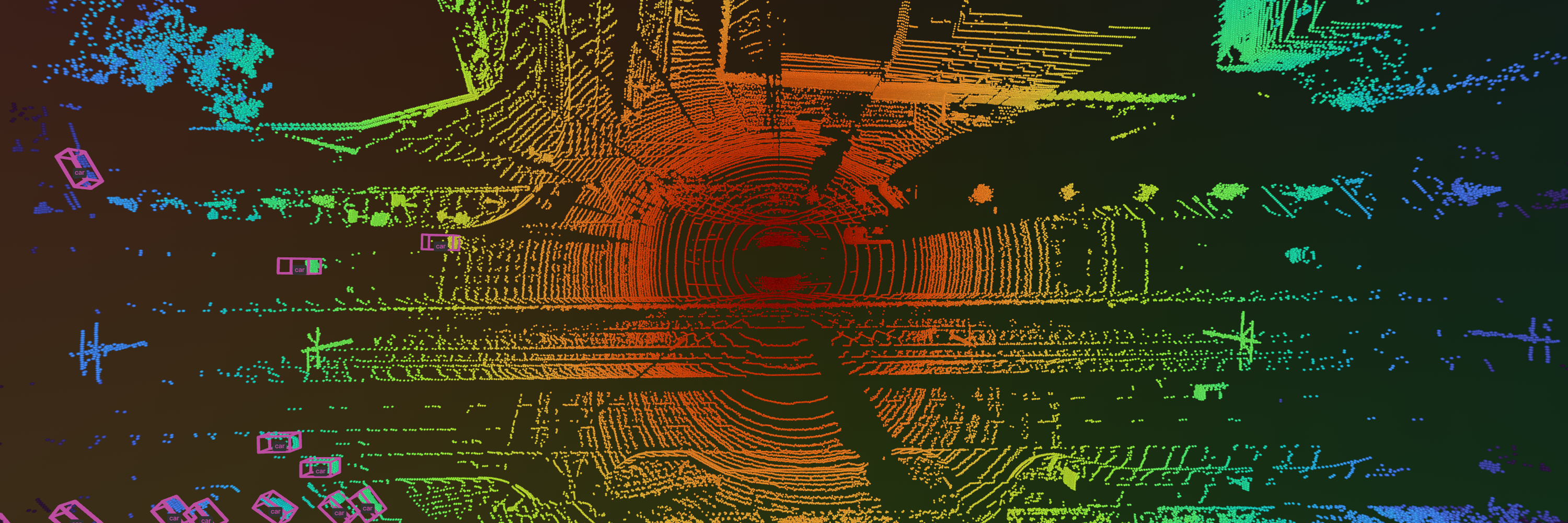}};
            \spy on (-0.013,0) in node [right] at (0.5,0.75);
            \spy on (-2.3,-0.9) in node [right] at (-3.7,0.75);
        \end{tikzpicture}
        \\

        \rotatebox{90}{\textbf{\hspace{0.12cm} \tiny ONCE (China)}} 
        &
        \begin{tikzpicture}[spy using outlines={circle,yellow,magnification=3,size=0.9cm, connect spies}]
            \node {\includegraphics[height=\heightof{\referencecamimage}, width=\widthof{\referencecamimage}]
                    {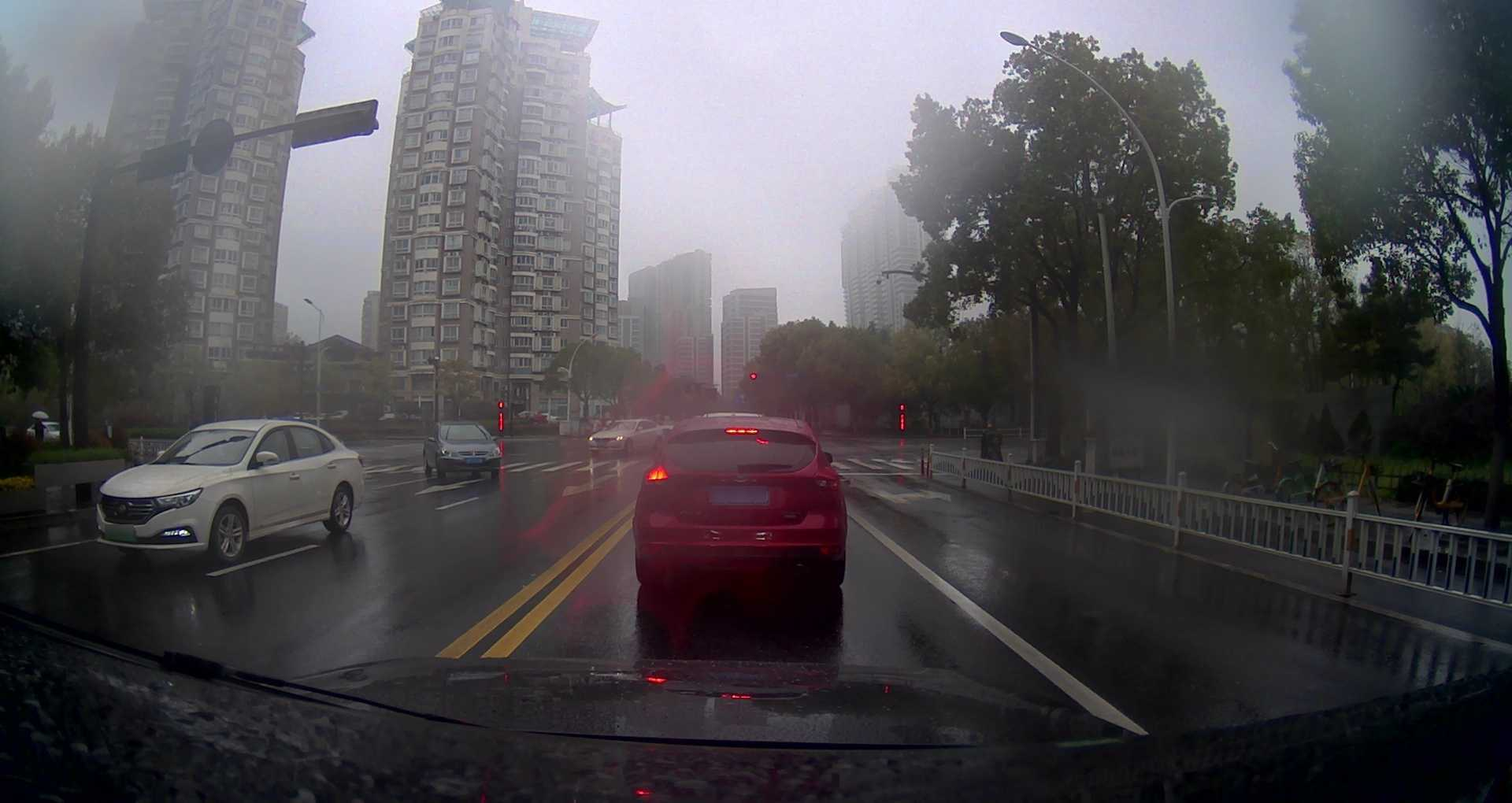}};
        \end{tikzpicture}
         &
        \begin{tikzpicture}[spy using outlines={circle,yellow,magnification=2,size=0.9cm, connect spies}]
            \node {\includegraphics[height=\heightof{\referencepcimage}, width=\widthof{\referencepcimage}]
                    {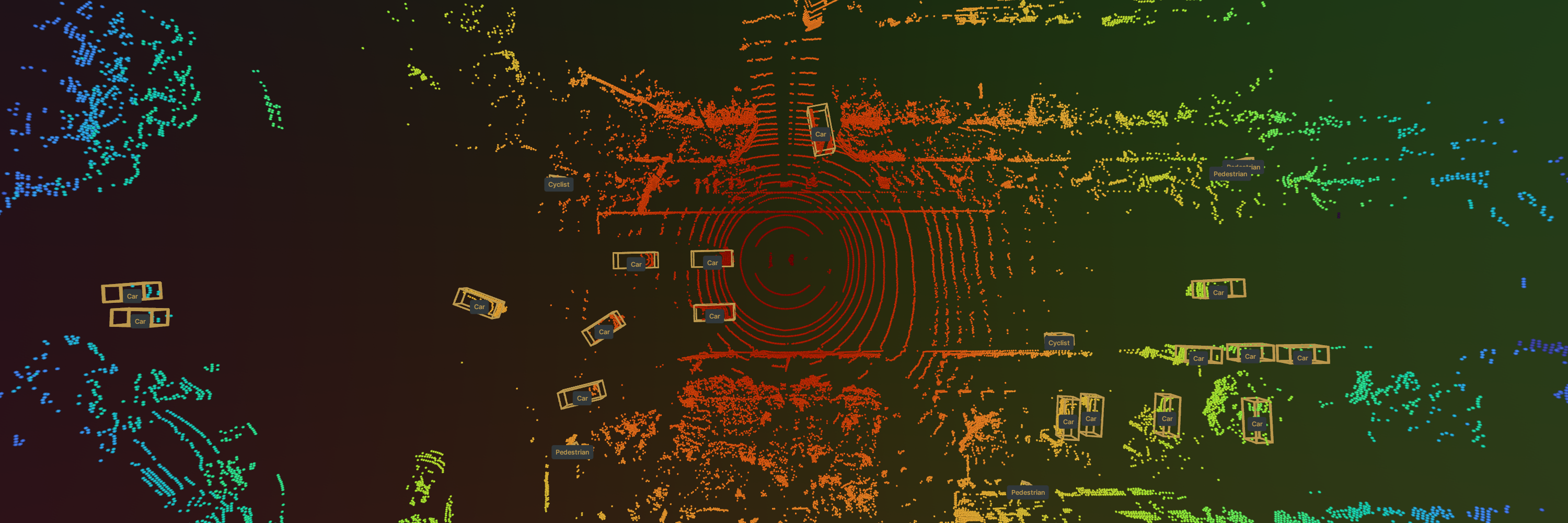} };
            \spy on (0.005,0) in node [right] at (0.6,0.75);
            \spy on (-3.1,-0.2) in node [right] at (-3.7,0.75);
        \end{tikzpicture} \\

    \end{tabular}
    \caption{\textbf{Datasets.} Exemplary databases captured in different ODDs. They utilize diverse LiDAR sensors, employ varied labeling specifications, and have different regions annotated. This diversity poses challenges in generalizing for 3D object detection models.}
    \label{fig:datasets}
\end{figure}

Most of these datasets, however, contain data gathered in only one city, state, country, or region (see Fig. \ref{fig:datasets}). As it was shown in \cite{train_in_germany}, an object detector trained on one dataset does not achieve comparable performance on another dataset due to regional differences in types of vehicles, traffic infrastructure, weather conditions, or fixed sensing apparatus,  showing very low model's generalization ability, especially for the LiDAR data.

It is the rule in the automotive industry that the operational limits of an automated driving system (ADS) are established early in the design process through the definition of its Operational Design Domain (ODD). The ODD specifies the conditions under which the ADS can safely operate and may be restricted to a specific region. However, the challenge arises when a company seeks to expand or adapt the ADS to a new ODD. In such cases, domain adaptation becomes a critical factor.

State-of-the-art object detectors trained on AV datasets usually exhibit poor performance when applied in regions or countries not represented in their training data. Deployment of these detectors is far from straightforward and requires adaptation to the new domain.
Such differences in the domain may be particularly visible in the case of cars, which differ substantially between Europe and the USA. The most popular car sold in Germany in 2023 was Volkswagen Golf \cite{golf_sales}, while in the United States Ford F-series reigns supreme \cite{f150_sales}. Comparing the lengths of the two vehicles -- 4.26m and 5.89m, respectively, one can note that the latter is 38\% longer than the former. 
This leads to scenarios where a car detector trained on the KITTI dataset (captured in Europe) achieves an average precision of 68.9\% when tested on KITTI but drops dramatically to just 12.3\% when evaluated on the Waymo dataset (recorded in the USA).

\begin{figure*}[!t]
    \centering
    \includegraphics[width=0.95\textwidth]{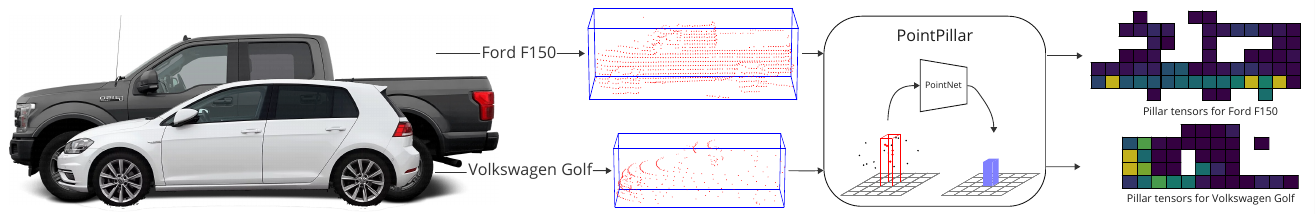}
    \caption{Vehicles of different sizes have different representations in the LiDAR point cloud, resulting in substantially different features being extracted.}
    \label{fig:feature_extraction}
\end{figure*}

 As can be seen in Fig. \ref{fig:feature_extraction}, the difference in vehicle size is reflected by object's representation in the LiDAR point cloud, which then affects their downstream representation in the most of the state-of-the-art 3D object detectors, which use voxelisation or other similar feature extraction techniques from LiDAR point cloud. The size and shape of the vehicle's point cloud representation significantly affects the feature vector extracted by the backbone of the 3D object detection models. For example, PointPillars \cite{pointpillar} extracts the features from LiDAR point cloud discretizing the 3D point cloud into a 2D pseudoimage, by dividing it into a set of pillars, which can be interpreted as a 2D grid in the bird's-eye view. Then, the collection of points from each pillar is encoded by a neural network which extracts the features from the pillar, returning a feature vector. We deeply investigate the chalenges in domain adaptation on various datasets in {\bf Appendices A and B}.

Domain adaptation may enable a model trained on one data distribution to perform effectively on a similar, yet distinct, distribution, even when limited target data are available. For instance, a detector trained on data from Germany can be adapted for deployment in the USA. This adaptation can be achieved either by retraining the model or by introducing a new model to transform the input data. In the context of autonomous driving, such techniques are highly valuable, as they minimize the need for collecting extensive, region-specific datasets.


Two general approaches to domain adaptation prevail in the literature. The first one is {\it fine-tuning} where the whole model is slightly adjusted to the few samples from the new distribution. The second one is {\it linear probing} where the source model freezes and only its detection head is exhaustively trained using the few new samples. In this paper, we consider domain adaptation in object detection for autonomous vehicles vision systems. We argue that in this application low-level features of the data are crucial for the system effectiveness. Therefore, we focus on fine-tuning and look for its best setting for this application.

In our approach to domain adaptation of autonomous driving data, the model is fine-tuned with some representative samples from the target domain. We adopt continual learning \cite{cl_and_cf} techniques to prevent the phenomenon of catastrophic forgetting \cite{cl_and_cf}, where learning new information by the model causes it to perform worse on older data.  In this work, we show that this approach delivers results comparable with the state-of-the-art domain adaptation algorithms, while also being less complicated and elaborate. 

The main contributions of this paper are:
\begin{itemize}
   \item We introduced a novel algorithm for LiDAR domain adaptation that uses neuron activation patterns to select the most diverse samples representing new domain.
    \item We prove that domain adaptation in the field of 3D object detection is feasible with only a few samples from the target domain.
    \item We evaluate several retraining and fine-tuning strategies to optimize model performance across both the source and target distributions comparing the results with state-of-the-art domain adaptation algorithms.

\end{itemize}

\section{Related work}

Domain adaptation for LiDAR data has gained significant attention in recent years due to the challenges posed by domain shifts in real-world applications. Over the years, various methods have been proposed to address these challenges.

{\bfseries Supervised domain adaptation} assumes that labeled data is available in both the source and target domains. Some solutions, such as \cite{wang2019range, rist2019cross, corral2021LiDAR, kim2024semi}, assume that target domain labels are pre-generated. To better generalize between domains, both \cite{wang2019range} and \cite{corral2021LiDAR} use discriminators to promote domain-invariant features. Additionally, \cite{corral2021LiDAR} adapts a 3D object detector to work within the CycleGAN \cite{zhu2017unpaired} framework. The method in \cite{kim2024semi} employs mixing augmentation accounting for LiDAR characteristics and point-wise adversarial augmentation to align the features in the target domain. The authors of \cite{rist2019cross} combine source datasets and propose an alternating training scheme across multiple tasks using a shared feature encoder.

Other approaches, such as \cite{yuan2023bi3d}, identify source-domain samples that closely resemble the target domain, maximizing performance gains when annotated. This strategy effectively reduces the annotation budget.

{\bfseries Unsupervised domain adaptation} aims to generalize a model to a target domain by using label information only from the source domain. The methods \cite{train_in_germany, luo2021unsupervised, yang2021st3d} focus on minimizing the geometric objects mismatch between the source and target domains. The authors of \cite{train_in_germany} apply a simple correction based on the average size of the vehicle. The network used in \cite{luo2021unsupervised} employs a teacher-student paradigm to generate reliable and adaptive pseudolabels. ST3D (Self-training for Unsupervised Domain Adaptation on 3D Object Detection) \cite{yang2021st3d} proposes to add random object scaling to pre-training on the source domain, which helps to overcome the bias in object size on the labeled source domain. A year later, the authors published ST3D++ \cite{yang2022st3d++}, which achieved even better results.

The authors of \cite{xu2021spg, wei2022lidar, saleh2019domain} indicate the difference in geometry of the point clouds as the main domain gap to overcome. SPG (Semantic Point Generation) \cite{xu2021spg} generates semantic points at the predicted foreground regions and faithfully recovers missing parts of the foreground objects. The method in \cite{wei2022lidar}, in each iteration, firstly generates low-beam pseudo LiDAR by downsampling the high-beam point clouds. Then the teacher-student framework is employed to distill rich information from the data with more beams. SRDAN (Scale-aware and Range-aware Domain Adaptation Network) \cite{zhang2021srdan} combines both scale-aware and range-aware domain alignments. The authors of \cite{saleh2019domain} employ a deep cycle-consistent generative adversarial network architecture to align synthetic and real point clouds in the bird's-eye view.

The methods \cite{saltori2020sf, fruhwirth2021fast3d, you2022exploiting} utilize object motion across sequential frames to create better pseudo-labels. SF-UDA$^{3D}$ (Source-Free Unsupervised Domain Adaptation) \cite{saltori2020sf}  uses the temporal consistency of detections between sequential frames. The authors of \cite{fruhwirth2021fast3d} utilize scene flow information to compute an object's mean flow, enhancing recall and enabling label propagation over time. Similarly, the method in \cite{you2022exploiting} generates pseudo-labels automatically after driving by replaying previously recorded sequences. Object tracks from these replays are smoothed forward and backward in time, with detections interpolated and extrapolated for improved accuracy.

Another approach is proposed in \cite{tsai2022see, lu2024dali}, which utilizes upsampled labels. SEE-MTDA \cite{tsai2022see} first isolates objects, recovers their geometry by surface completion and then upsample every object’s triangle mesh to increase the confidence of 3D detectors. DALI \cite{lu2024dali} reduces bias in pseudo-label size distribution through a post-training size normalization strategy. To mitigate instance-level noise, it generates pseudo-points using a combination of a 3D model library and a LiDAR sensor library.
CMT (Co-training Mean-Teacher) \cite{chen2024cmt} constructs a hybrid domain that aligns domain-specific features more effectively. The authors also developed batch adaptive normalization to achieve a more stable learning process.


\section{Proposed domain adaptation}
\label{sec:method}


{\bf \noindent Formal problem definition.}
We consider the problem of domain adaptation. In its {\it basic} version, it is defined as follows. We are given a neural model, $M_{\wei_0}$ with weights $\wei_0$, fitted to a dataset, $D_0$, sampled from a distribution, $P_0$. The problem is to adjust the model $M_{\wei_0}$ to fit a distribution $P_1$, which is similar to $P_0$ but still different. To this end, a dataset $D_1$ is given, sampled from $P_1$. However, $D_1$ is too small to support retraining $M$ without significant overfitting. 

We are also interested in three additional variations of the above basic problem. The {\it second} considered problem is to adjust the model $M_{\wei_0}$ to fit both $P_0$ and $P_1$ at the same time. The {\it third} considered problem is to train the model $M_{\wei_1}$ with the limited $K$ datasets $D_1,\dots,D_K$ for the model to fit the distributions $P_1,\dots,P_K$. The {\it fourth} problem is to train the single model $M_{\wei_1}$ with the limited datasets $D_1,\dots,D_K$, for the model $M_{\wei_1}$ to fit all the distributions $P_0,P_1,\dots,P_K$. 

{\bf \noindent Interpretations.}
In this paper, we focus on a specific example of the above basic problem. That is, we consider $M_{\wei_0}$ as a detector of 3D objects in LiDAR images taken from a car. The model has been trained on a dataset recorded in a certain geographic region. We wish to adjust the model to detect objects in a different geographic region. To this end, we have a handful of samples recorded in this other region. 

Secondly, we still require the new model $M_{\wei_1}$ to work well also in the original region. Thirdly, we require the trained model to work well in several other regions from which we have a handful of training samples. Fourthly, we require the model to behave universally well both in the original region and in multiple others.  

{\bf \noindent Approaches.}
\label{sec:approaches}
Looking for a solution to the above problems, we notice that the model $M_{\wei_0}$ has already been trained to identify all features of the data necessary to process samples from both the $P_0$ and $P_1$ distributions. However, the same features may manifest slightly differently in $P_1$. 

Therefore, in this paper, we verify several approaches to the domain adaptation problem. They are all based on {\it post-training}, which is training a model $M_\wei$ to fit the given samples $D_1$, while keeping $\wei$ close to $\wei_0$. 

We consider several strategies of post-training. In all strategies, we commence the post-training from $\wei=\wei_0$. Next, one of the following is realized: 
\begin{itemize}
  \item \textbf{L2-SP} regularization \cite{xuhong2018explicit} -- during post-training, a penalty, $\Omega$, is added to the loss function for deviating from the pre-trained model weights. The penalty function $\Omega$ is of the form 
          $\Omega(\bm{w}) = \alpha {\lVert \bm{w} - \bm{w}_0 \rVert}^2,$
        where $\bm{\alpha}$~is the regularization parameter setting the strength of the penalty.
  \item \textbf{Linear fading} of learning rate -- during post-training, we turn off the Adam Onecycle optimiser and replace it with a learning rate that linearly fades with each epoch to $0$.
  \item \textbf{Small constant} learning rate -- we turn off the Adam Onecycle optimiser and replace it with a small constant learning rate. Here, the resulting weights $\wei$ are close to $\wei_0$ just because both the learning rate and the number of post-training epochs are small. 
\end{itemize} 
In our experiments, for comparison, we also apply standard post-training techniques:
\begin{itemize} 
  \item \textbf{Usual fine-tuning} -- the post-training takes a limited number of epochs and its setting is similar to that applied in the original training. 
  \item \textbf{Linear probing}  -- in this approach only the last linear layer of the model is updated and the rest of the weights in the model are frozen.
\end{itemize}

{
\section{Source-Target Distribution Alignment}

Source--target distribution matching has become a standard component of recent semi-supervised and unsupervised domain adaptation pipelines for 3D object detection. In our approach, we adopt the following two widely used techniques.

\subsection{Point Cloud Density Alignment}

A major source of domain shift arises from differences in LiDAR point cloud density across datasets, caused by variations in sensor configurations, scanning patterns, and scene layouts. Several recent studies have demonstrated that aligning point cloud density between source and target domains during pretraining is highly effective. State-of-the-art methods such as LiDAR Distillation \cite{wei2022lidar}, SSDA3D \cite{ssda3d}, TODA \cite{toda3d}, and Bi3D \cite{yuan2023bi3d} consistently show that pretraining on source point clouds that are downsampled or upsampled to match the target-domain density leads to substantial performance improvements after adaptation.

This strategy reduces geometric discrepancies between domains and enables the model to learn representations that are less sensitive to sensor-specific sampling characteristics. As a result, LiDAR distribution matching has become a de facto standard procedure in modern 3D domain adaptation and semi-supervised learning pipelines.

\subsection{Bounding Box Size Alignment}

In addition to point cloud density, object size distributions may differ significantly across datasets. Vehicle dimensions, for example, vary across regions and datasets due to differences in traffic composition and annotation standards. The work of \cite{train_in_germany} introduced statistical normalization (SN, also referred to as Stat-Norm) to address this issue by aligning 3D bounding box dimension statistics between source and target domains.

Stat-Norm rescales source-domain bounding boxes to match the target-domain size distribution, thereby reducing scale-related biases during training. This simple yet effective technique has been shown to significantly improve cross-domain generalization and is now widely adopted in LiDAR-based domain adaptation methods.

}

{
\section{Diverse Frame Selection Algorithm}

To efficiently select informative target domain frames for fine-tuning, we propose a \emph{diverse frame selection algorithm} based on latent activation patterns in the PV-RCNN ROI Head.

\subsection{Activation Pattern Extraction}

We extract \emph{activation patterns} from ReLU-activated convolutional layers as follows:

\textbf{Feature Attribution:} Each 3D bounding box $b$ is associated with its latent ReLU-activated vector $\mathbf{f}_b \in \mathbb{R}^d$ produced by the selected ROI Head layer.
\textbf{Clipping:} We zero out the lesser half of the vector values, keeping only the larger, more informative components:
\[
\mathbf{f}_b^{\text{clip}} = \text{clip}_{\text{top-half}}(\mathbf{f}_b)
\]
\textbf{Binarization:} The remaining positive components are set to 1 to form a \emph{binary activation pattern}:
\[
\mathbf{p}_b = \text{sign}(\mathbf{f}_b^{\text{clip}}), \quad \mathbf{p}_b \in \{0,1\}^d
\]

This transforms latent features into discrete patterns encoding the presence of strong activations for each bounding box.

\subsection{Layer Selection for Pattern Extraction}

To select the most informative layer, we leverage the labeled \emph{source domain validation set}. For each candidate layer, we compute activation patterns for true positive (TP) and false positive (FP) boxes detected on the source validation split. For each box $b$, we compute the Hamming distance to the nearest neighbor in the bank of source domain ground-truth patterns:
\[
D(b) = \min_{\mathbf{p}_g \in \mathcal{P}_{\text{GT}}} \text{Hamming}(\mathbf{p}_b, \mathbf{p}_g),
\]
where $\mathcal{P}_{\text{GT}}$ is the set of all GT patterns from the source domain.

We then evaluate how well the layer separates TP and FP boxes using AUROC over the nearest neighbor distances. The selected layer is the one for which TP patterns are closest to GT patterns while FP patterns are farthest. Our experiments show that the first ROI Head layer yields the most informative activation patterns.

\subsection{Frame-Level Diversity Score}

Given a target-domain frame $f$ with detected boxes $\mathcal{B}(f)$, each 
box $b \in \mathcal{B}(f)$ is assigned an integer-valued similarity score
$D(b)$ equal to the Hamming distance to its nearest source GT pattern.

Since Hamming distances are integers, the set of distances
\[
\mathcal{D}_f = \{D(b) : b \in \mathcal{B}(f)\}
\]
induces a discrete probability distribution without any need for binning.  
Let $\pi_f(k)$ denote the normalized frequency of distance value $k$:

\[
\pi_f(k)
= \frac{1}{|\mathcal{B}(f)|}
\left|
\{ b \in \mathcal{B}(f) : D(b) = k \}
\right|,
\qquad 
\sum_{k} \pi_f(k) = 1.
\]

This distribution characterizes the heterogeneity of detected objects in latent 
space. Frames spanning a wide range of similarity values (i.e., containing both 
source-like and highly out-of-distribution objects) exhibit higher entropy.

We therefore define the entropy score
\[
H(f) = 
- \sum_{k} \pi_f(k)\, \log \pi_f(k),
\]
which serves as an intra-frame diversity measure.

\subsection{Iterative Diverse Frame Selection}

To prevent selecting multiple frames with similar characteristics, we adopt an 
iterative strategy in which the top-$K$ proposal set is recomputed at every 
selection step. This ensures that once a frame is selected, it is removed from 
future candidate pools.

Let $\mathcal{F}_{\text{target}}$ be the set of all target-domain frames and 
$\mathcal{F}_{\text{sel}}^{(t)}$ the set of frames selected after iteration $t$.

\begin{enumerate}
    \item Construct the iteration-specific proposal set:
    \[
        \mathcal{F}_K^{(t)}
        = \mathrm{TopK}\!\left(H(f),\;
        f \in \mathcal{F}_{\text{target}} \setminus \mathcal{F}_{\text{sel}}^{(t-1)}
        \right).
    \]

    \item For each $f \in \mathcal{F}_K^{(t)}$ compute the inter-frame distance
    using a compact average-over-pairs form:
    \[
        \mathrm{Dist}(f,\mathcal{F}_{\text{sel}}^{(t-1)})
        =
        \begin{cases}
            1, & \mathcal{F}_{\text{sel}}^{(t-1)}=\emptyset, \\[6pt]
            \dfrac{1}{|\mathcal{F}_{\text{sel}}^{(t-1)}|}
            \!\sum\limits_{f' \in \mathcal{F}_{\text{sel}}^{(t-1)}}
            \overline{\mathrm{Ham}}(f,f'),
            & \text{otherwise},
        \end{cases}
    \]
    where the mean pairwise Hamming distance between two frames is
    \[
        \overline{\mathrm{Ham}}(f,f')
        = \frac{1}{|\mathcal{B}(f)|\,|\mathcal{B}(f')|}
          \sum_{b \in \mathcal{B}(f)} \sum_{b' \in \mathcal{B}(f')}
          \mathrm{Ham}(\mathbf{p}_b,\mathbf{p}_{b'}).
    \]

    \item Normalize both diversity factors over the proposal set 
    \( \mathcal{F}_K^{(t)} \):
    \[
        \hat{H}(f) = \mathrm{norm}\!\left(H(f)\right),
    \]
    \[
        \widehat{\mathrm{Dist}}(f)
        = \mathrm{norm}\!\left(\mathrm{Dist}(f,\mathcal{F}_{\text{sel}}^{(t-1)})\right).
    \]

    \item Select the next frame:
    \[
        f^* = \arg\max_{f \in \mathcal{F}_K^{(t)}}
              \hat{H}(f)\cdot\widehat{\mathrm{Dist}}(f),
    \]
    and update
    \[
        \mathcal{F}_{\text{sel}}^{(t)}
        = \mathcal{F}_{\text{sel}}^{(t-1)} \cup \{f^*\}.
    \]

    \item Repeat until $|\mathcal{F}_{\text{sel}}^{(t)}| = N$.
\end{enumerate}

This process guarantees that each iteration considers only the most informative 
yet-unselected frames, while also encouraging selection diversity at both the 
frame and detection levels.

\begin{algorithm}[t]
\caption{Diverse Frame Selection for Target Domain}
\label{alg:diverse_frames}
\begin{algorithmic}[1]

\Require Pretrained source detector, target frames $\mathcal{F}_{\text{target}}$, 
         proposal size $K$, desired number $N$
\State Extract source GT activation patterns $\mathcal{P}_{\text{GT}}$
\State Infer all target frames; collect activation patterns $\mathbf{p}_b$ for each box
\For{each frame $f \in \mathcal{F}_{\text{target}}$}
    \State Compute similarity scores $D(b)$ to $\mathcal{P}_{\text{GT}}$
    \State Compute entropy $H(f)$ from discrete distribution $\pi_f(k)$
\EndFor
\State $\mathcal{F}_{\text{sel}}^{(0)} \gets \emptyset$, \quad $t \gets 1$
\While{$|\mathcal{F}_{\text{sel}}^{(t-1)}| < N$}
    \State $\mathcal{F}_K^{(t)} \gets \mathrm{TopK}\big(H(f),\, f \in \mathcal{F}_{\text{target}} \setminus \mathcal{F}_{\text{sel}}^{(t-1)}\big)$
    \For{each $f \in \mathcal{F}_K^{(t)}$}
        \State Compute $\mathrm{Dist}(f,\mathcal{F}_{\text{sel}}^{(t-1)})$ (or set to $1$ if empty)
    \EndFor
    \State $\hat{H}(f) \gets \mathrm{norm}(H(f))$ over $\mathcal{F}_K^{(t)}$
    \State $\widehat{\mathrm{Dist}}(f) \gets \mathrm{norm}(\mathrm{Dist}(f,\mathcal{F}_{\text{sel}}^{(t-1)}))$ over $\mathcal{F}_K^{(t)}$
    \State $f^* \gets \arg\max_{f \in \mathcal{F}_K^{(t)}} \hat{H}(f)\cdot\widehat{\mathrm{Dist}}(f)$
    \State $\mathcal{F}_{\text{sel}}^{(t)} \gets \mathcal{F}_{\text{sel}}^{(t-1)} \cup \{f^*\}$
    \State $t \gets t + 1$
\EndWhile
\State \Return $\mathcal{F}_{\text{sel}}^{(t-1)}$
\end{algorithmic}
\end{algorithm}

}

\section{Experiments and Analysis}

\subsection{Setup}
{\bf \noindent Datasets.} For the purposes of our study, we used the KITTI dataset \cite{kitti_dataset}, the NuScenes \cite{nuscenes_dataset} and Waymo \cite{nuscenes_dataset} datasets converted to the KITTI dataset format. We converted the datasets using the code attached to \cite{train_in_germany}. Additionally, in the converted point clouds, we scaled the range of the intensity attribute of the points to the range $[0.0, 1.0]$, which is the same as in the data from the KITTI dataset.

We also created smaller subsets of each dataset with 10 and 100 frames from the training set. These were used in experiments simulating limited target data, ensuring a diverse range of object sizes, distances, and LiDAR viewing angles.

{\bf \noindent 3D object detection.} For 3D object detection in point clouds, we used TED~\cite{TED}, a method that achieves state-of-the-art performance on the KITTI benchmark.
TED employs a sparse convolutional backbone to extract voxel features. Subsequently, these features are aligned and aggregated into compact representations, enabling 3D object detection. Additionally, we use PV-RCNN \cite{shi2020pvrcnn} architecture because this model is widely used in the state-of-the-art papers related to domain adaptation problem. PV-RCNN uses 3D voxel convolutional neural network and PointNet-based set abstraction to learn more discriminative point cloud features.

In our experiments, we trained the model in $40$ epochs with a starting learning rate parameter equal to $0.01$ set during training by Adam optimizer \cite{smith2019super}. We focused only on  {\it{Car}} class because this category was also reported in state-of-the-art papers.


{\bf \noindent Metrics.} We evaluated 3D object detection performance using standard methods and metrics commonly applied to object detection algorithms on the KITTI dataset. In the results, we reported the values of the metric $AP_{3D}$ (Average Precision) at IoU = \{0.5, 0.7\} using the same protocol as it was used in \cite{train_in_germany}.

\subsection{Results of domain adaptation}\label{sec:baseline_exp}
We conducted a series of experiments to evaluate the proposed domain adaptation strategies, assessing the model’s performance after adaptation to identify the most effective approach. 
To ensure a comprehensive comparison, we benchmarked the best-performing adaptation methods against state-of-the-art domain adaptation techniques. 

It is important to note that our results may differ from those obtained on the same datasets without applying the dataset conversion method proposed in \cite{train_in_germany}, even when using the same algorithm. This discrepancy arises due to differences in data preprocessing across datasets. Notably, the performance on nuScenes is significantly lower than the results reported in the official nuScenes's benchmark. This is likely due to the relative sparsity of the point cloud in nuScenes compared to other datasets, as well as differences in preprocessing—particularly the filtering of ground truth boxes in KITTI based on the number of points inside, which reduces the number of training examples.

{\bf \noindent Size of the target domain.}
In the first experiment, we examined the impact of the number of target domain samples used in post-training. In this experiment, we address the basic problem of domain adaptation with the usual fine-tuning (Sec.~\ref{sec:method}). The results are contained in Tab.~\ref{tab:10_100_post-training}. Starting with models trained on the source dataset, we evaluated their performance on the validation sets of all three datasets. We denote as {{source dataset}} the dataset on which the model was trained, and as {{target dataset}} the one on which it was evaluated. 
To assess the effect of post-training, we fine-tuned each pre-trained model using small subsets of the target dataset. 
We used 10- and 100-frame subsets of the target dataset, selected either randomly or to maximize object diversity by including vehicles at varying distances from the ego vehicle.

\begin{table*}[htbp]
\centering
\caption{Impact of post-training on the model performance using varying frames from target dataset.}
\label{tab:10_100_post-training}
\scriptsize
\begin{tabular}{cccccccc}
\toprule
\multirow{2}{*}{\diagbox{\textbf{Src.}}{\textbf{Tar.}}} & \textbf{Number} & \multicolumn{2}{c}{\textbf{KITTI}} & \multicolumn{2}{c}{\textbf{NuScenes}} & \multicolumn{2}{c}{\textbf{Waymo}} \\
\cmidrule(lr){3-4} \cmidrule(lr){5-6} \cmidrule(lr){7-8}
& \textbf{of frames} & $\bm{AP_{0.5}}$ & $\bm{AP_{0.7}}$ & $\bm{AP_{0.5}}$ & $\bm{AP_{0.7}}$ & $\bm{AP_{0.5}}$ & $\bm{AP_{0.7}}$ \\
\midrule
\multirow{5}{*}{\textbf{KITTI}} 
& 0 & 96.80 & 87.99 & 23.91 & 7.26 & 52.92 & 11.21 \\
& +10 (div.) & -- & -- & 28.40 & 14.35 & 67.02 & 51.55 \\
& +100 (div.) & -- & -- & 33.89 & 17.35 & 72.60 & 58.22 \\
& +10 (rand.) & -- & -- & 27.49 & 13.18 & 66.62 & 48.17 \\
& +100 (rand.) & -- & -- & 34.21 & 17.92 & 72.16 & 57.10 \\
\midrule
\multirow{5}{*}{\textbf{NuScenes}} 
& 0 & 80.57 & 23.58 & 42.32 & 25.96 & 44.46 & 18.58 \\
& +10 (div.) & 94.34 & 81.09 & -- & -- & 66.14 & 49.04 \\
& +100 (div.) & 96.82 & 82.45 & -- & -- & 72.80 & 57.60 \\
& +10 (rand.) & 92.71 & 79.23 & -- & -- & 65.51 & 47.58 \\
& +100 (rand.) & 96.89 & 83.16 & -- & -- & 72.42 & 57.65 \\
\midrule
\multirow{5}{*}{\textbf{Waymo}} 
& 0 & 88.93 & 7.68 & 19.40 & 5.39 & 78.70 & 67.17 \\
& +10 (div.) & 95.07 & 81.95 & 27.07 & 10.98 & -- & -- \\
& +100 (div.) & 97.39 & 85.14 & 34.56 & 17.64 & -- & -- \\
& +10 (rand.) & 94.15 & 80.87 & 27.91 & 11.29 & -- & -- \\
& +100 (rand.) & 97.26 & 85.40 & 34.44 & 17.38 & -- & -- \\
\bottomrule
\end{tabular}
\end{table*}

\begin{figure}[!htbp]
  \centering
    \includegraphics[width=0.75\linewidth]{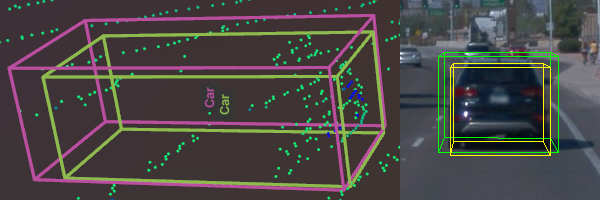}

    \includegraphics[width=0.75\linewidth]{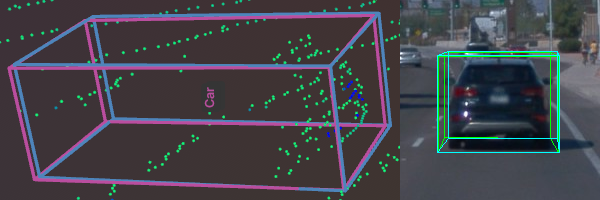}

  \caption{Exemplary object detection in new domain (top) without and  with  (bottom) domain adaptation.}
  \label{fig:object_detection}

\end{figure}

It is evident that post-training on a limited subset of the data results in a notable enhancement, while the annotation budget required to annotate 10 or 100 samples is very low. We can also observe that for 10 samples diverse object's selection is better than random one. This leads to a reduction in the discrepancy between the AP metric value calculated with IoU thresholds of 0.5 and 0.7. Consequently, it enables for a more precise estimation of the object's position. Fig. \ref{fig:object_detection} illustrates the detection of the same object by a model that did not undergo post-training and by a model that did. 
We also conducted an additional experiment to assess the effectiveness of training the model from scratch on a selected subset (See {\bf Appendix C}).

{\bf \noindent Post-training on single vs. multiple domains.}
In the second experiment, we investigated whether domain adaptation could be applied simultaneously to multiple domains and what is the impact on the source domain, addressing four problems of Sec.~\ref{sec:method} with the usual fine-tuning. We used a combined 10-frame subsets of the KITTI, NuScenes and Waymo datasets. We ran post-training on the models learned on each dataset and evaluations of each model resulting from post-training using the validation sets of all datasets. We summarised the results in Tab. \ref{tab:post_KNW}.
\begin{table*}[htbp]
\centering
\caption{Post-training impact on the model performance using only 10 frames from single or multiple target domains.}
\label{tab:post_KNW}
\scriptsize
\begin{tabular}{cccccccc}
\toprule
\multirow{2}{*}{\diagbox{\textbf{Src.}}{\textbf{Tar.}}} & \multirow{2}{*}{\textbf{Post-training}} & \multicolumn{2}{c}{\textbf{KITTI}} & \multicolumn{2}{c}{\textbf{NuScenes}} & \multicolumn{2}{c}{\textbf{Waymo}} \\
\cmidrule(lr){3-4} \cmidrule(lr){5-6} \cmidrule(lr){7-8}
& & $\bm{AP_{0.5}}$ & $\bm{AP_{0.7}}$ & $\bm{AP_{0.5}}$ & $\bm{AP_{0.7}}$ & $\bm{AP_{0.5}}$ & $\bm{AP_{0.7}}$ \\
\midrule
\multirow{3}{*}{\textbf{KITTI}} 
& w/o & 96.80 & 87.99 & 23.91 & 7.26 & 52.92 & 11.21 \\
& target & -- & -- & 28.25 & 14.19 & 66.77 & 46.04 \\
& multi & 93.72 & 83.03 & 28.62 & 13.88 & 60.56 & 29.89 \\
\midrule
\multirow{3}{*}{\textbf{NuScenes}} 
& w/o & 80.57 & 23.58 & 42.32 & 25.96 & 44.46 & 18.58 \\
& target & 93.45 & 79.86 & -- & -- & 65.64 & 49.40 \\
& multi & 93.45 & 79.24 & 30.69 & 13.43 & 58.25 & 30.69 \\
\midrule
\multirow{3}{*}{\textbf{Waymo}} 
& w/o & 88.93 & 7.68 & 19.40 & 5.39 & 78.70 & 67.17 \\
& target & 96.73 & 83.03 & 30.86 & 12.25 & -- & -- \\
& multi & 93.57 & 81.21 & 28.97 & 14.35 & 64.29 & 35.28 \\
\bottomrule
\end{tabular}
\end{table*}

The results obtained from post-training on data from different datasets are superior to those obtained on datasets other than the source dataset. However, the improvement observed is consistently less pronounced than that which would be achieved if only the target dataset were used. Furthermore, the post-training process has a markedly detrimental impact on the results obtained on the source dataset, indicating that it is not possible to develop a universal model through this approach. These results indicate that when a company plans to extend the ODD across multiple regions, domain adaptation must be performed separately for each region.  Additionally, we tested the model’s performance on the original domain after adaptation (See {\bf Appendix C}).

{\bf \noindent Post-training strategies.} In the third experiment, we evaluated different post-training strategies, checking the impact of the changes in learning rate and regularisation method, using various approaches proposed in Sec.~\ref{sec:method}.
The optimal parameters were identified as a penalty factor for the L2-SP regularisation method of 0.01 or 0.001, contingent on the adaptation task; a starting learning rate for the linear learning rate fade of 0.01 or 0.005; and a learning rate of 0.001 or 0.005 for post-training with a constant learning rate.
The results obtained using the optimal parameters are presented in Table~\ref{tab:method_post-training_TED}, providing a comparison between the baseline performance without post-training, the results after applying post-training with default parameters, and the outcomes achieved using the proposed methods. We also evaluated the impact of post-training strategy on pedestrian class (see {\bf Appendix C}).

\begin{table*}[htbp]
\centering
\caption{Domain adaptation results for different post-training methods on 10 frames from the target domain.}
\label{tab:method_post-training_TED}
\scriptsize
\begin{tabular}{cccccccc}
\toprule
\multirow{2}{*}{\diagbox{\textbf{Src.}}{\textbf{Tar.}}} & \multirow{2}{*}{\textbf{Post-training}} & \multicolumn{2}{c}{\textbf{KITTI}} & \multicolumn{2}{c}{\textbf{NuScenes}} & \multicolumn{2}{c}{\textbf{Waymo}} \\
\cmidrule(lr){3-4} \cmidrule(lr){5-6} \cmidrule(lr){7-8}
& & $\bm{AP_{0.5}}$ & $\bm{AP_{0.7}}$ & $\bm{AP_{0.5}}$ & $\bm{AP_{0.7}}$ & $\bm{AP_{0.5}}$ & $\bm{AP_{0.7}}$ \\
\midrule
\multirow{6}{*}{\textbf{KITTI}} 
& without & 96.80 & 87.99 & 23.91 & 7.26 & 52.92 & 11.21 \\
& default & -- & -- & 28.40 & 14.35 & 67.02 & 51.55 \\
& L2-SP reg. & -- & -- & 33.94 & 20.16 & 68.69 & 51.59 \\
& LR fading & -- & -- & 33.81 & 19.85 & 67.57 & 51.21 \\
& const LR & -- & -- & 33.05 & 19.90 & 67.42 & 50.72 \\
& linear probing & -- & -- & 30.04 & 16.00 & 61.78 & 40.84 \\
\midrule
\multirow{6}{*}{\textbf{NuScenes}} 
& without & 80.57 & 23.58 & 42.32 & 25.96 & 44.46 & 18.58 \\
& default & 94.34 & 81.09 & -- & -- & 66.14 & 49.04 \\
& L2-SP reg. & 94.57 & 82.10 & -- & -- & 68.26 & 51.70 \\
& LR fading & 94.36 & 81.86 & -- & -- & 68.82 & 51.64 \\
& const LR & 95.56 & 81.31 & -- & -- & 67.37 & 50.33 \\
& linear probing & 94.85 & 79.16 & -- & -- & 65.44 & 47.55 \\
\midrule
\multirow{6}{*}{\textbf{Waymo}} 
& without & 88.93 & 7.68 & 19.40 & 5.39 & 78.70 & 67.17 \\
& default & 95.07 & 81.95 & 27.07 & 10.98 & -- & -- \\
& L2-SP reg. & 97.15 & 85.01 & 34.47 & 15.10 & -- & -- \\
& LR fading & 96.65 & 84.76 & 32.97 & 14.34 & -- & -- \\
& const LR & 96.83 & 84.17 & 34.70 & 15.92 & -- & -- \\
& linear probing & 94.82 & 76.39 & 33.78 & 16.18 & -- & -- \\
\bottomrule
\end{tabular}
\end{table*}

The results show that selection of the correct post-training strategy has the potential to significantly enhance the adaptation performance. The proposed methods are better than linear probing and in most scenarios L2-SP Regularisation achieves the best performance, however the adaptation score may vary between different datasets. An alternative approach to optimising the adaptation score could be the imposition of constraints on the post-training phase, aimed at preventing the model from deviating significantly from the original weights. 

{\bf \noindent Comparison with state-of-the-art.} The final experiment evaluated whether the proposed domain adaptation strategy, using only 10 representative frames from the target domain with different post-training strategy, outperforms state-of-the-art methods. To ensure a fair comparison, we used the same PV-RCNN architecture \cite{shi2020pvrcnn} as in prior state-of-the-art papers related to domain adaptation problem. The results, presented in Tab. \ref{tab:compare} and \ref{tab:compare50}, show that the proposed approach with post-training using only 10 frames, combined with some post-training strategy, delivers surprisingly good performance, surpassing state-of-the-art domain adaptation techniques. Moreover, this performance can be even higher when evaluation is performed following the evaluation protocol presented in \cite{tsai2022see}. Our approach was outperformed only by Bi3D \cite{yuan2023bi3d} in the Waymo $\rightarrow$ nuScenes adaptation scenario, however Bi3D utilized 1\% of the target dataset for adaptation, which corresponds to 281 frames. 

\begin{table}[!htbp]
 \caption{ Comparison with state-of-the art domain adaptation techniques and Source Only adaptation.}

  \scriptsize
  \centering
  \label{tab:compare}
  \tabcolsep=0.11cm
  \begin{tabular}{|c|c|c|c|}
    \hline
    \textbf{Task}               & \textbf{Method}   & $\bm{AP_{0.7}}$                        & \textbf{$\Delta\bm{AP_{0.7}}$} \\
       \hline\hline
        \multirow{9}{*}{\textbf{N. $\rightarrow$ K.}}  & Target domain                       & 84.66                                 & --                              \\
                                                                & Source only                         & 37.26                                 & --                              \\
                                                                & Stat-Norm \cite{train_in_germany}   & 49.47                                 & +12.21  \\
                                                                & ST3D \cite{yang2021st3d}            & 70.85                                 & +33.59  \\
                                                                & Bi3D (281 f.) \cite{yuan2023bi3d}      & 77.55                                 & +40.29  \\
                                                                & CMT \cite{chen2024cmt}              & 75.51                                 & +38.25  \\                   

& \textbf{10 f. + const LR (ours)} & \textbf{78.17} & \textbf{+40.91} \\

& \textbf{37 f. + const LR (ours)} & \textbf{80.21} & \textbf{+42.95} \\

& \textbf{100 f. + const LR (ours)} & \textbf{80.90} & \textbf{+43.64} \\
        \hline
        \multirow{9}{*}{\textbf{W. $\rightarrow$ K.}}    & Target domain                       & 84.66                                 & --                              \\
                                                                & Source only                         & 29.24                                 & --                              \\
                                                                & Stat-Norm \cite{train_in_germany}   & 63.60                                 & +34.36  \\
                                                                & ST3D \cite{yang2021st3d}            & 64.78                                 & +35.54  \\
                                                                & Bi3D (281 f.) \cite{yuan2023bi3d}      & 78.03                                 & +48.79  \\
                                                                & CMT \cite{chen2024cmt}              & 74.53                                 & +45.29  \\

& \textbf{10 f. + const LR (ours)} & \textbf{79.22} & \textbf{+49.98} \\

& \textbf{37 f. + const LR (ours)} & \textbf{81.39} & \textbf{+52.15} \\

& \textbf{100 f. + const LR (ours)} & \textbf{82.23} & \textbf{+52.99} \\                                                                
                                                                \hline

    \multirow{ 10}{*}{\textbf{W. $\rightarrow$ N.}} & Target domain                       & 35.51                                 & --           \\
                                                             & Source only                         & 19.16                                  & --         \\
                                                             & Stat-Norm \cite{train_in_germany}   & 22.29                                 & +3.13     \\
                                                             & ST3D \cite{yang2021st3d}            & 22.99                                 & +3.83     \\
                                                             & {Bi3D (281 f.) \cite{yuan2023bi3d}  }    & {  30.81}                                 & {+11.65 }     \\
                                                             & CMT \cite{chen2024cmt}              & 26.38                                 & +7.22         \\

& 10 f. + const LR (ours)         & 29.57 & +10.41 \\

& 100 f. + const LR (ours)        & 32.14 & +12.98 \\

&  \textbf{281 f. + const LR (ours)}  & \textbf{33.41} & \textbf{+14.25} \\

& SSDA3D (281 f.) \cite{ssda3d}         & 33.01 & +13.85 \\

& TODA (281 f.) \cite{toda3d}         & 32.18 & +13.02 \\

                                                             \hline

  \end{tabular}
\end{table}

\begin{table}
 \caption{Evaluation results on KITTI-format val set with ground truth 3D bounding boxes having <50 points being filtered out as in SEE \cite{tsai2022see}.}

  \scriptsize
  \centering
  \label{tab:compare50}
  \tabcolsep=0.11cm
  \begin{tabular}{|c|c|c|c|}
    \hline
    \rowcolor{white}
\textbf{Adaptation task}               & \textbf{Method}   & $\bm{AP_{0.7}}$                        & \textbf{$\Delta\bm{AP_{0.7}}$} \\ \hline
        \multirow{6}{*}{\textbf{NuScenes $\rightarrow$ KITTI}} & Target domain                       & 91.19                                 & --                   \\
                                                                      & Source only                         & 45.27                                 & --                   \\
                                                                      & SEE \cite{tsai2022see}              & 72.51                                 & +27.24  \\
                                                                      & 10 f. + const LR (ours)             & 82.12                                 & +36.85  \\
                                                                      & 10 f. + LR fading (ours)            & 81.76                                 & +36.49  \\
                                                                      & {\bf 10 f. + L2-SP (ours)}          & {\bf 85.56}                           & {\bf +40.29}  \\
        \hline
        \multirow{6}{*}{\textbf{Waymo $\rightarrow$ KITTI}}    & Target domain                       & 91.19                                 & --                   \\
                                                                      & Source only                         & 33.71                                 & --                   \\
                                                                      & SEE \cite{tsai2022see}              & 79.39                                 & +45.68  \\
                                                                      & 10 f. + const LR (ours)             & 88.10                                 & +54.39  \\
                                                                      & 10 f. + LR fading (ours)            & 87.53                                 & +53.82  \\
                                                                      & {\bf 10 f. + L2-SP (ours)}          & {\bf 88.91}                           & {\bf +55.20}  \\
    \hline
        \multirow{6}{*}{\textbf{Waymo $\rightarrow$ NuScenes}}    & Target domain                       & 70.56                                 & --                   \\
                                                                      & Source only                         & 9.01                                 & --                   \\
                                                                      & SEE \cite{tsai2022see}              & 14.72                                 & +5.71  \\
                                                                      & 10 f. + const LR (ours)             & 53.57                                 & +44.56  \\
                                                                      & 10 f. + LR fading (ours)            & 44.01                                 & +35.00  \\
                                                                      & {\bf 10 f. + L2-SP (ours)}          & {\bf 56.14}                           & {\bf +47.13}  \\
    \hline
  \end{tabular}
\end{table}





\begin{table}
\caption{ Ablation study for FT+SN - target domain N frames finetuning with 3D bounding box statistical normalization, LiDAR distribution alignment for pretraining, and activation-pattern entropy frame selection. }
\centering
\scriptsize
\renewcommand{\arraystretch}{1.15}
{
\begin{tabular}{|c|c|c|c|c|}
\hline
\textbf{Frames} 
& \textbf{FT + SN} 
& \textbf{LIDAR dist. align.} 
& \textbf{NAP}
& \textbf{Waymo→NuScenes} \\
\hline

\multirow{4}{*}{10}
&  &  &   & 19.16 (source only) \\
& + &  &   & 24.71 ( +5.55 ) \\
& + & + &   & 27.49 ( +2.78 )\\
& + & + & + & 29.57 \,(\,+2.08\,) \\
\hline

\multirow{4}{*}{100}
&  &  &  & 19.16 (source only) \\
& + &  &  & 27.62 ( +8.46 ) \\
& + & + &  & 31.68 ( +4.06 )\\
& + & + & + & 32.14 \,(\,+0.46\,) \\
\hline

\multirow{4}{*}{281}
&  &  &   & 19.16 (source only) \\
& + &  &   & 28.29 ( +9.13 )\\
& + & + &   & 32.93 ( +4.64 )\\
& + & + & + & 33.41 \,(\,+0.48\,) \\
\hline
\end{tabular}
}
\end{table}

With just 10 annotated, representative samples that requires only a few minutes of manual effort, the proposed method achieves performance comparable to training on the full target dataset. However, performance declines when linear probing is used as the post-training strategy. These results indicate that while state-of-the-art object detectors excel at feature extraction, accurate object localization still benefits greatly from domain adaptation. The key challenge lies in selecting a small yet representative subset from the target domain. These findings highlight an important implication: autonomous driving companies aiming to extend the ODD to new regions or sensor configurations can build a primary training dataset and then apply domain adaptation using only a small number of annotated samples from the new domain.

These findings highlight a crucial insight. When autonomous driving companies seek to extend the ODD to new regions or different sensor setups, they can only prepare a small training database and then apply domain adaptation with a significantly smaller set of annotated samples from the new domain. This may save the cost of expensive data collection and labeling in new domains.

\section{Conclusion}
In this paper, we presented that simple domain adaptation techniques based on a handful of training examples and post-training strategies preventing against weight drift from the original model allows to achieve much better results than linear probing and other state-of-the-art domain adaptation techniques for 3D object detection in the context of autonomous driving. Domain adaptation is crucial when OEMs seek to expand the ODD, as the conventional approach typically involves collecting a new training dataset from the target domain. Our experiments demonstrated that domain adaptation can be a more efficient alternative, achieving performance levels close to those obtained with a fully trained model on the target domain. 
In addition, domain adaptation across multiple domains is less effective than adaptation to a single domain and after domain adaptation the model's performance declines on the source dataset. These findings demonstrate that extending the ODD in autonomous driving applications can be achieved with only a limited number of target domain samples. 

\backmatter

\section*{Acknowledgements}
This work was supported by the National Centre for Research and Development under the research project \mbox{GOSPOSTRATEG8/0001/2022}.

\bibliography{bibliography}

\begin{thebibliography}{48}
\providecommand{\natexlab}[1]{#1}
\providecommand{\url}[1]{{#1}}
\providecommand{\urlprefix}{URL }
\providecommand{\doi}[1]{\url{https://doi.org/#1}}
\providecommand{\eprint}[2][]{\url{#2}}
 \bibcommenthead

\bibitem[{Alibeigi et~al.(2023)Alibeigi, Ljungbergh, Tonderski et~al.}]{zenseact_dataset}
Alibeigi M, Ljungbergh W, Tonderski A, et~al (2023) Zenseact open dataset: A large-scale and diverse multimodal dataset for autonomous driving. In: Proc. of the Int. Conf. on Computer Vision ({ICCV}), pp 20178--20188

\bibitem[{Bekker(2024)}]{golf_sales}
Bekker H (2024) Best selling cars in {Germany} in 2023. \url{https://www.best-selling-cars.com/germany/2023-full-year-germany-best-selling-car-models/}, accessed: 2024-09-20

\bibitem[{Caesar et~al.(2020)Caesar, Bankiti, Lang et~al.}]{nuscenes_dataset}
Caesar H, Bankiti V, Lang AH, et~al (2020) nuscenes: A multimodal dataset for autonomous driving. In: Proc. of the Conference on Computer Vision and Pattern Recognition ({CVPR}), pp 11621--11631

\bibitem[{Carballo et~al.(2020)Carballo, Lambert, Monrroy et~al.}]{9304681}
Carballo A, Lambert J, Monrroy A, et~al (2020) {LIBRE: The Multiple 3D LiDAR Dataset}. In: 2020 {IEEE} Intelligent Vehicles Symposium (IV), pp 1094--1101, \doi{10.1109/IV47402.2020.9304681}

\bibitem[{Chen et~al.(2024)Chen, Zhuo, Li et~al.}]{chen2024cmt}
Chen S, Zhuo J, Li X, et~al (2024) {CMT: Co-training Mean-Teacher for Unsupervised Domain Adaptation on 3D Object Detection}. In: {ACM} Multimedia 2024

\bibitem[{Corral-Soto et~al.(2021)Corral-Soto, Nabatchian, Gerdzhev et~al.}]{corral2021LiDAR}
Corral-Soto ER, Nabatchian A, Gerdzhev M, et~al (2021) Lidar few-shot domain adaptation via integrated cyclegan and {3D} object detector with joint learning delay. In: Proc. of the Int. Conf. on Robotics and Automation ({ICRA}), IEEE, pp 13099--13105

\bibitem[{Fang et~al.(2024)Fang, Zhou, Zhao et~al.}]{fang2024LiDAR}
Fang J, Zhou D, Zhao J, et~al (2024) {LiDAR-CS dataset: LiDAR point cloud dataset with cross-sensors for 3D object detection}. In: Proc. of the Int. Conf. on Robotics and Automation ({ICRA}), IEEE, pp 14822--14829

\bibitem[{Fruhwirth-Reisinger et~al.(2021)Fruhwirth-Reisinger, Opitz, Possegger et~al.}]{fruhwirth2021fast3d}
Fruhwirth-Reisinger C, Opitz M, Possegger H, et~al (2021) {FAST3D}: Flow-aware self-training for {3D} object detectors. arXiv preprint arXiv:211009355

\bibitem[{Geiger et~al.(2012)Geiger, Lenz, and Urtasun}]{kitti_dataset}
Geiger A, Lenz P, Urtasun R (2012) Are we ready for {Autonomous} {Driving?} {The} {KITTI} {Vision} {Benchmark} {Suite}. In: Proc. of the Conference on Computer Vision and Pattern Recognition ({CVPR})

\bibitem[{Geyer et~al.(2020)Geyer, Kassahun, Mahmudi et~al.}]{geyer2020a2d2}
Geyer J, Kassahun Y, Mahmudi M, et~al (2020) A2d2: Audi autonomous driving dataset. arXiv preprint arXiv:200406320

\bibitem[{Heinzler(2022)}]{heinzler2022LiDAR}
Heinzler RK (2022) Lidar-based weather detection: Automotive lidar sensors in adverse weather conditions

\bibitem[{Houston et~al.(2021)Houston, Zuidhof, Bergamini et~al.}]{lyft_dataset}
Houston J, Zuidhof G, Bergamini L, et~al (2021) One thousand and one hours: Self-driving motion prediction dataset. In: Proc. of the Conference on Robot Learning ({CORL}), PMLR, pp 409--418

\bibitem[{Kim et~al.(2024{\natexlab{a}})Kim, Lee, Park, Kim, Lim, Chang, and Choi}]{toda3d}
Kim Y, Lee J, Park C, et~al (2024{\natexlab{a}}) Semi-supervised domain adaptation using target-oriented domain augmentation for 3d object detection. {IEEE} Transactions on Intelligent Vehicles

\bibitem[{Kim et~al.(2024{\natexlab{b}})Kim, Lee, Park et~al.}]{kim2024semi}
Kim Y, Lee J, Park C, et~al (2024{\natexlab{b}}) Semi-supervised domain adaptation using target-oriented domain augmentation for {3D} object detection. {IEEE} Trans on Intelligent Vehicles

\bibitem[{Lang et~al.(2019)Lang, Vora, Caesar, Zhou et~al.}]{pointpillar}
Lang AH, Vora S, Caesar H, et~al (2019) {Pointpillars: Fast encoders for object detection from point clouds}. In: Proc. of the Conference on Computer Vision and Pattern Recognition ({CVPR}), pp 12697--12705

\bibitem[{Li and Ibanez-Guzman(2020)}]{9127855}
Li Y, Ibanez-Guzman J (2020) {Lidar for Autonomous Driving: The Principles, Challenges, and Trends for Automotive Lidar and Perception Systems}. {IEEE} Signal Processing Magazine 37(4):50--61. \doi{10.1109/MSP.2020.2973615}

\bibitem[{Lu and Radha(2024)}]{lu2024dali}
Lu X, Radha H (2024) {DALI: Domain Adaptive LiDAR Object Detection via Distribution-level and Instance-level Pseudo Label Denoising}. {IEEE} Trans on Robotics

\bibitem[{Luo et~al.(2021)Luo, Cai, Zhou et~al.}]{luo2021unsupervised}
Luo Z, Cai Z, Zhou C, et~al (2021) Unsupervised domain adaptive {3D} detection with multi-level consistency. In: Proc. of the Int. Conf. on Computer Vision ({ICCV}), pp 8866--8875

\bibitem[{Mao et~al.(2021)Mao, Niu, Jiang et~al.}]{mao2021one}
Mao J, Niu M, Jiang C, et~al (2021) {One Million Scenes for Autonomous Driving: ONCE Dataset}. arXiv preprint arXiv:210611037

\bibitem[{Matuszka et~al.(2022)Matuszka, Barton, Butykai et~al.}]{Matuszka2022aiMotiveDA}
Matuszka T, Barton I, Butykai {\'A}, et~al (2022) {aiMotive} {Dataset}: A {Multimodal} {Dataset} for {Robust} {Autonomous} {Driving} with {Long-Range} {Perception}. ArXiv abs/2211.09445. \urlprefix\url{https://api.semanticscholar.org/CorpusID:253581598}

\bibitem[{Pham et~al.(2020)Pham, Sevestre, Pahwa et~al.}]{9197385}
Pham QH, Sevestre P, Pahwa RS, et~al (2020) {A*3D Dataset: Towards Autonomous Driving in Challenging Environments}. In: Proc. of the Int. Conf. on Robotics and Automation ({ICRA}), pp 2267--2273, \doi{10.1109/ICRA40945.2020.9197385}

\bibitem[{Rist et~al.(2019)Rist, Enzweiler, and Gavrila}]{rist2019cross}
Rist CB, Enzweiler M, Gavrila DM (2019) Cross-sensor deep domain adaptation for lidar detection and segmentation. In: Proc. of the Intelligent Vehicles Symposium (IV), IEEE, pp 1535--1542

\bibitem[{Saleh et~al.(2019)Saleh, Abobakr, Attia et~al.}]{saleh2019domain}
Saleh K, Abobakr A, Attia M, et~al (2019) Domain adaptation for vehicle detection from bird's eye view {LiDAR} point cloud data. In: Proc. of the Int. Conf. on Computer Vision Workshops ({ICCV}), pp 0--0

\bibitem[{Saltori et~al.(2020)Saltori, Lathuili{\'e}re, Sebe et~al.}]{saltori2020sf}
Saltori C, Lathuili{\'e}re S, Sebe N, et~al (2020) {SF-UDA}$^{3D}$: Source-free unsupervised domain adaptation for {LiDAR}-based {3D} object detection. In: Proc. of the Int. Conf. on 3D Vision (3DV), IEEE, pp 771--780

\bibitem[{Sanchez et~al.(2024)Sanchez, Soum-Fontez, Deschaud et~al.}]{10508037}
Sanchez J, Soum-Fontez L, Deschaud JE, et~al (2024) {ParisLuco3D:} a {High-Quality} {Target} {Dataset} for {Domain} {Generalization} of {LiDAR} {Perception}. {IEEE} Robotics and Automation Letters 9(6):5496--5503. \doi{10.1109/LRA.2024.3393209}

\bibitem[{Shi et~al.(2020)Shi, Guo, Jiang, Wang, Shi, Wang, and Li}]{shi2020pvrcnn}
Shi S, Guo C, Jiang L, et~al (2020) Pv-rcnn: Point-voxel feature set abstraction for 3d object detection. In: Proc. of the Conference on Computer Vision and Pattern Recognition ({CVPR}), pp 10526--10535

\bibitem[{Smith(2024)}]{f150_sales}
Smith C (2024) Best selling cars in the {USA} in 2023. \url{https://www.motor1.com/features/703891/best-selling-cars-2023/}, accessed: 2024-09-20

\bibitem[{Smith and Topin(2019)}]{smith2019super}
Smith LN, Topin N (2019) Super-convergence: Very fast training of neural networks using large learning rates. In: Artificial intelligence and machine learning for multi-domain operations applications, SPIE, pp 369--386

\bibitem[{Sun et~al.(2020)Sun, Kretzschmar, Dotiwalla et~al.}]{waymo_dataset}
Sun P, Kretzschmar H, Dotiwalla X, et~al (2020) Scalability in perception for autonomous driving: Waymo open dataset. In: Proc. of the Conference on Computer Vision and Pattern Recognition ({CVPR})

\bibitem[{Tsai et~al.(2022)Tsai, Berrio, Shan et~al.}]{tsai2022see}
Tsai D, Berrio JS, Shan M, et~al (2022) See eye to eye: A lidar-agnostic {3D} detection framework for unsupervised multi-target domain adaptation. {IEEE} Robotics and Automation Letters 7(3):7904--7911

\bibitem[{van~de Ven et~al.(2024)van~de Ven, Soures, and Kudithipudi}]{cl_and_cf}
van~de Ven GM, Soures N, Kudithipudi D (2024) Continual learning and catastrophic forgetting. \urlprefix\url{https://arxiv.org/abs/2403.05175}, {\href{https://arxiv.org/abs/2403.05175}{{arXiv:2403.05175}}}

\bibitem[{Wang et~al.(2018)Wang, Huang, Cheng et~al.}]{Wang2018TheAO}
Wang P, Huang X, Cheng X, et~al (2018) The {ApolloScape} {Open} {Dataset} for {Autonomous} {Driving} and {Its} {Application}. {IEEE} Trans on Pattern Analysis and Machine Intelligence 42:2702--2719. \urlprefix\url{https://api.semanticscholar.org/CorpusID:52842854}

\bibitem[{Wang et~al.(2020)Wang, Chen, You et~al.}]{train_in_germany}
Wang Y, Chen X, You Y, et~al (2020) {Train in Germany, Test in the USA: Making 3D Object Detectors Generalize}. In: Proc. of the Conference on Computer Vision and Pattern Recognition ({CVPR}), pp 11710--11720

\bibitem[{Wang et~al.(2023)Wang, Yin, Li, Frossard, Yang, and Shen}]{ssda3d}
Wang Y, Yin J, Li W, et~al (2023) Ssda3d: Semi-supervised domain adaptation for 3d object detection from point cloud. In: Proceedings of the 37th {AAAI} Conference on Artificial Intelligence ({AAAI}), pp 2707--2715

\bibitem[{Wang et~al.(2019)Wang, Ding, Li et~al.}]{wang2019range}
Wang Z, Ding S, Li Y, et~al (2019) Range adaptation for {3D} object detection in {LiDAR}. In: Proc. of the Int. Conf. on Computer Vision Workshops ({ICCV}), pp 0--0

\bibitem[{Wei et~al.(2022)Wei, Wei, Rao et~al.}]{wei2022lidar}
Wei Y, Wei Z, Rao Y, et~al (2022) {LiDAR Distillation: Bridging the Beam-Induced Domain Gap for {3D} Object Detection}. In: Proc. of the European Conference on Computer Vision ({ECCV}), Springer, pp 179--195

\bibitem[{Wilson et~al.(2021)Wilson, Qi, Agarwal et~al.}]{argoverse_dataset}
Wilson B, Qi W, Agarwal T, et~al (2021) {Argoverse 2: Next Generation Datasets for Self-Driving Perception and Forecasting}. In: Proc. of the Neural Information Processing Systems Track on Datasets and Benchmarks (NeurIPS Datasets and Benchmarks 2021)

\bibitem[{Wu et~al.(2023)Wu, Wen, Li et~al.}]{TED}
Wu H, Wen C, Li W, et~al (2023) {Transformation-Equivariant 3D Object Detection for Autonomous Driving}. In: {AAAI}

\bibitem[{Xiao et~al.(2021)Xiao, Shao, Hao et~al.}]{xiao2021pandaset}
Xiao P, Shao Z, Hao S, et~al (2021) Pandaset: Advanced sensor suite dataset for autonomous driving. In: Proc. of the International Intelligent Transportation Systems Conference ({ITSC}), IEEE, pp 3095--3101

\bibitem[{Xu et~al.(2021)Xu, Zhou, Wang et~al.}]{xu2021spg}
Xu Q, Zhou Y, Wang W, et~al (2021) {SPG}: Unsupervised domain adaptation for {3D} object detection via semantic point generation. In: Proc. of the Int. Conf. on Computer Vision ({ICCV}), pp 15446--15456

\bibitem[{Xuhong et~al.(2018)Xuhong, Grandvalet, and Davoine}]{xuhong2018explicit}
Xuhong L, Grandvalet Y, Davoine F (2018) Explicit inductive bias for transfer learning with convolutional networks. In: Proc. of the Int. Conf. on Machine Learning ({ICML}), PMLR, pp 2825--2834

\bibitem[{Yang et~al.(2021)Yang, Shi, Wang et~al.}]{yang2021st3d}
Yang J, Shi S, Wang Z, et~al (2021) {ST3D}: Self-training for unsupervised domain adaptation on {3D} object detection. In: Proc. of the Int. Conf. on Computer Vision and Pattern Recognition ({CVPR}), pp 10368--10378

\bibitem[{Yang et~al.(2022)Yang, Shi, Wang et~al.}]{yang2022st3d++}
Yang J, Shi S, Wang Z, et~al (2022) {ST3D++}: Denoised self-training for unsupervised domain adaptation on {3D} object detection. {IEEE} Trans on pattern analysis and machine intelligence 45(5):6354--6371

\bibitem[{You et~al.(2022)You, Diaz-Ruiz, Wang et~al.}]{you2022exploiting}
You Y, Diaz-Ruiz CA, Wang Y, et~al (2022) Exploiting playbacks in unsupervised domain adaptation for {3D} object detection in self-driving cars. In: Proc. of the Int. Conf. on Robotics and Automation ({ICRA}), IEEE, pp 5070--5077

\bibitem[{Yuan et~al.(2023)Yuan, Zhang, Yan et~al.}]{yuan2023bi3d}
Yuan J, Zhang B, Yan X, et~al (2023) {Bi3D}: Bi-domain active learning for cross-domain {3D} object detection. In: Proc. of the Conference on Computer Vision and Pattern Recognition ({CVPR}), pp 15599--15608

\bibitem[{Zhang et~al.(2023)Zhang, Yuan, Shi et~al.}]{Zhang_2023_CVPR}
Zhang B, Yuan J, Shi B, et~al (2023) {Uni3D: A Unified Baseline for Multi-Dataset 3D Object Detection}. In: Proc. of the Conference on Computer Vision and Pattern Recognition ({CVPR}), pp 9253--9262

\bibitem[{Zhang et~al.(2021)Zhang, Li, and Xu}]{zhang2021srdan}
Zhang W, Li W, Xu D (2021) {SRDAN}: Scale-aware and range-aware domain adaptation network for cross-dataset {3D} object detection. In: Proc. of the Conference on Computer Vision and Pattern Recognition ({CVPR}), pp 6769--6779

\bibitem[{Zhu et~al.(2017)Zhu, Park, Isola et~al.}]{zhu2017unpaired}
Zhu JY, Park T, Isola P, et~al (2017) Unpaired image-to-image translation using cycle-consistent adversarial networks. In: Proc. of the Int. Conf. on Computer Vision ({ICCV}), pp 2223--2232

\end{thebibliography}

\newpage




\begin{appendices}

\section{Challenges in domain adaptation}
Effective LiDAR domain adaptation requires a comprehensive understanding of the characteristics of the LiDAR point clouds inherent to various datasets and the environmental conditions under which they were collected. This section examines how sensor properties and environmental factors influence model performance in domain adaptation. Additionally, we provide an overview of prominent LiDAR datasets, detailing their sensor specifications and discuss how the domain adaptation can be affected. The summary of the factors that can affect domain adaptation between regions or environmental conditions is presented in Tab. \ref{tab:datasets}. The goal of this analysis is to show when domain adaptation could be easier, which can support the decision about recording a new training database, when a product is planned to be deployed in a new environment, e.g., city or country or with a new LiDAR sensor.

\begin{figure}[phtb]
    \centering
    \includegraphics[width=\linewidth]{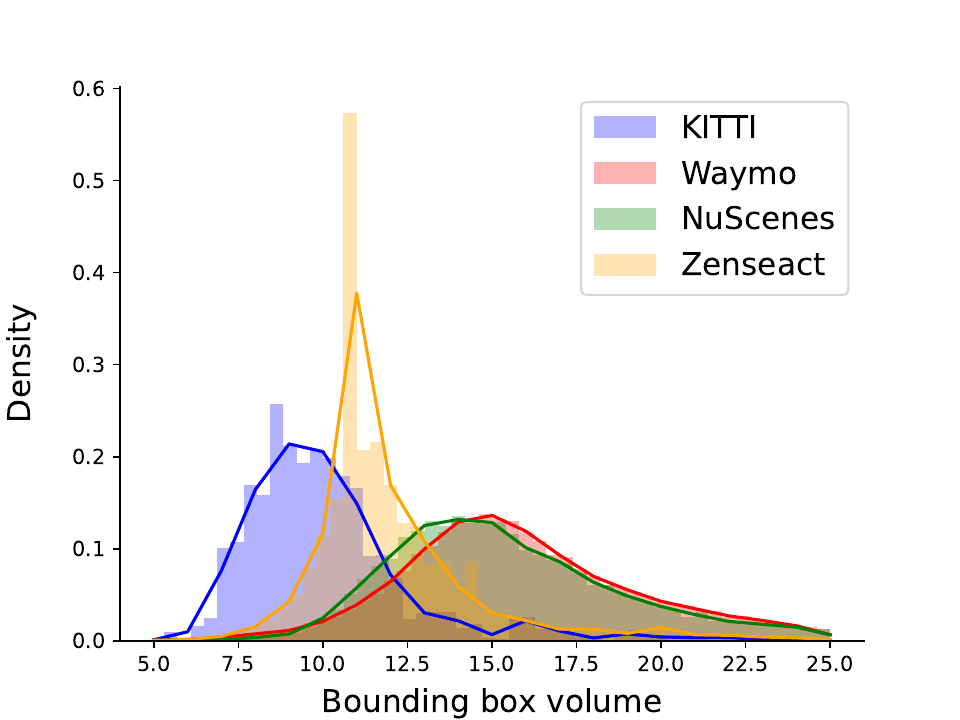}\\
    \caption{Distribution of volumes of vehicle annotations. These distributions result from the cultural and geographical features as well as from the labeling guideline.}
    \label{fig:vehicle_size_datasets}
\end{figure}

\subsection{Sensor Characteristics}
\subsubsection{Resolution and Point Cloud Density}
Lidar sensors differ in resolution, affecting point cloud density. High-resolution sensors produce dense point clouds, while low-resolution ones result in sparser data. This disparity presents challenges when transferring models between datasets. In case of domain transfer from higher to lower resolution, we can directly remove the redundant LiDAR lines, but adaptation from lower to higher LiDAR resolution is more challenging.

Horizontal and vertical resolution in LiDAR sensors differ due to the underlying scanning patterns and sensor configurations. Horizontal resolution is generally determined by the angular increment between successive laser beams as the sensor rotates around its vertical axis (mechanical spinning LiDARs). Since a full rotation typically covers 360°, a finer horizontal resolution means more laser pulses per rotation, leading to a denser horizontal sampling of the environment. In contrast, vertical resolution is defined by the number and spacing of the laser channels in the sensor stacked vertically. A LiDAR sensor with more vertical channels or tighter vertical spacing produces denser sampling in the vertical direction.

Because of these differing resolutions, the same object within a scene may be represented by varying numbers of points both horizontally and vertically. A sensor with high vertical resolution can capture a more detailed structure of the height and shape of an object, providing a rich vertical point distribution. Meanwhile, a sensor with higher horizontal resolution will present a more detailed view of the width and lateral features of an object. When transitioning between datasets with different horizontal and vertical resolutions, models must effectively adapt to these variations in point density, ensuring that features remain robustly identifiable even when the underlying point distributions along horizontal or vertical axes differ significantly.

\begin{figure}[phtb]
    \centering
    \includegraphics[width=\linewidth]{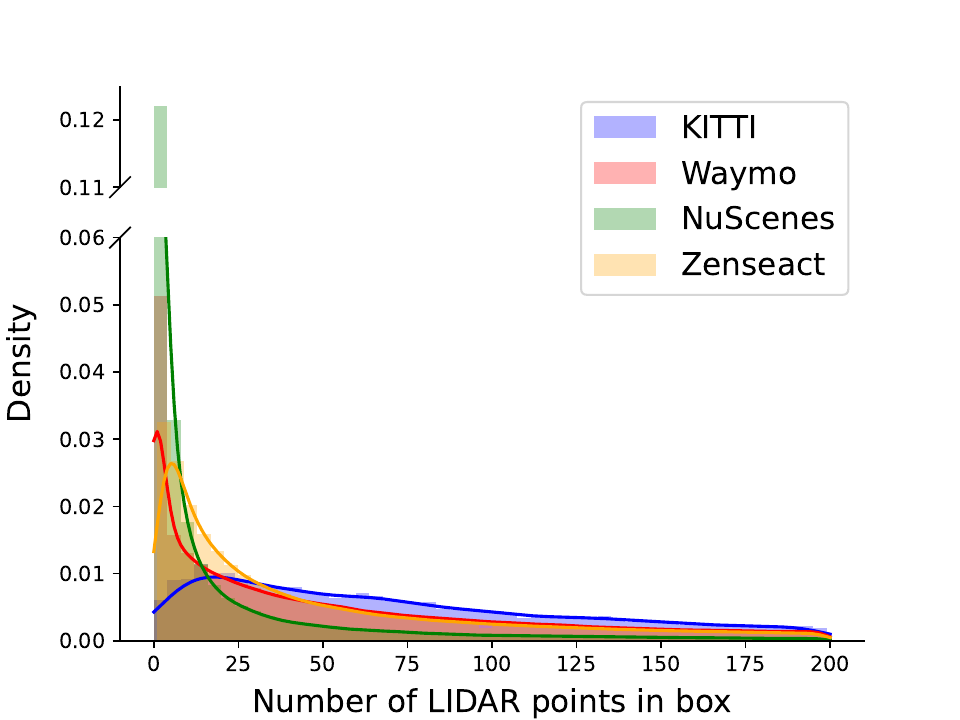}\\
    \caption{Distribution of LiDAR points in the annotation boxes. These distributions result from the point cloud density and the labeling policy}
    \label{fig:num_pts_datasets}
\end{figure}

For example, direct transfer of models between datasets with differing sensor resolutions, such as from nuScenes (32 beams) to KITTI (64 beams), yields notable performance reductions. This outcome is not surprising when observing the distribution of points per object presented in Figure \ref{fig:num_pts_datasets}. Domain adaptation techniques such as ST3D \cite{yang2021st3d} and ST3D++ \cite{yang2022st3d++} have demonstrated notable improvements, particularly in the context of adapting from lower-resolution to higher-resolution datasets.

\subsubsection{Field of View (FoV)}
Differences in vertical and horizontal FoV influence the extent of the captured environment. For example, the Waymo Open Dataset, which uses multiple LiDARs with varying FoVs, provides extensive environmental coverage. A narrow FoV can lead to missing information, complicating model transfer across datasets.

Furthermore, some datasets (e.g. original KITTI or Zenseact Open Dataset) contain annotations for the objects only visible from the front-facing camera. The cameras also may have different FoVs - the one from KITTI has horizontal resolution of 90° while the camera used in Zenseact captures 120° of the environment. Thus, the models trained on such datasets will better detect objects located in front of the ego vehicle than those located by the sides or behind it.

Finally, the translation and rotation of the sensors must be considered. The same object captured by the same LiDAR will appear differently from different viewpoints. In general, the objective is to maximize the number of points representing objects of interest. Therefore, when creating such a dataset, it is worthwhile to test different sensors' position setups to ensure optimal results.

\subsubsection{Beam Pattern and Scanning Mechanism}
Variations in beam patterns (e.g., uniform vs. non-uniform) and scanning mechanisms (mechanical vs. solid-state) affect point distribution. For example, the Hesai PandarGT sensor used in PandaSet, has a different scanning mechanism compared to the Velodyne sensors used in the most popular open datasets, potentially leading to inconsistencies in model transfer.

Mechanical spinning systems, such as those used in popular Velodyne sensors, use rotating mirrors to achieve the 360° horizontal FoV, but are bulky and susceptible to mechanical wear in harsh conditions. In contrast, solid-state systems, including MEMS-based LiDARs, reduce moving parts by utilizing miniature electromechanical mirrors, offering improved robustness for automotive use. Flash LiDARs, a type of solid-state system, completely eliminate moving parts by illuminating the entire scene simultaneously with a single pulse and capturing reflections with a 2D sensor array. However, they are limited in range because of power restrictions for eye safety and their inability to dynamically adjust the FoV.

Newer optical phased-array (OPA) technologies provide true solid-state scanning by steering laser beams through phase modulation, but their commercial adoption remains limited. Each of these mechanisms impacts the density and uniformity of point cloud data. According to \cite{9127855}, the selection of the appropriate scanning mechanism is critical to adapt LiDAR systems to specific operational environments, as differences in beam density, range, and precision influence perception performance.

\begin{figure*}[ht]
    \centering
    
    \begin{subfigure}[b]{0.3\textwidth}
        \centering
        \includegraphics[width=\textwidth]{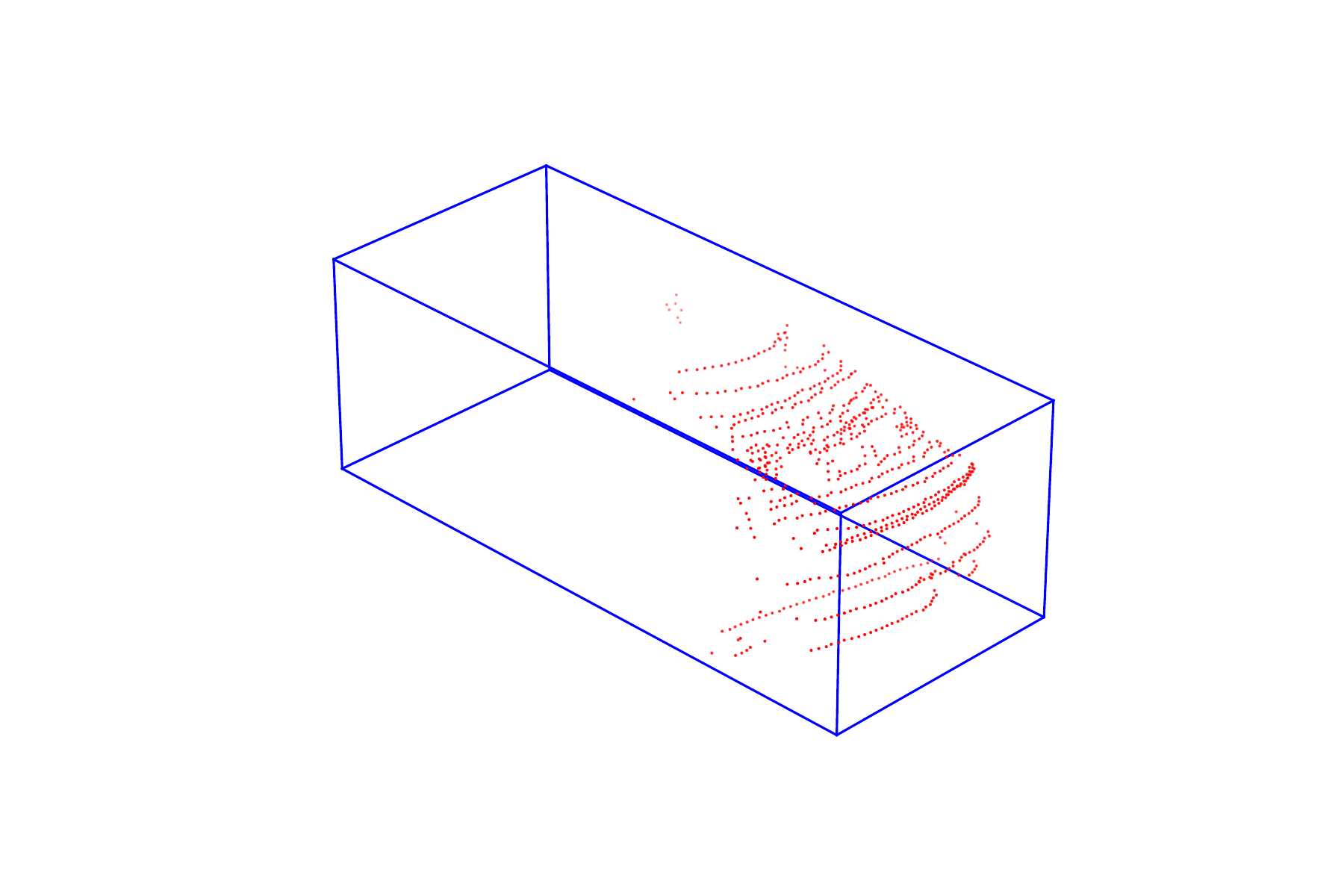}
        \caption{Car point cloud - mechanical spinning LiDAR}
    \end{subfigure}
    \hfill
    \begin{subfigure}[b]{0.3\textwidth}
        \centering
        \includegraphics[width=\textwidth]{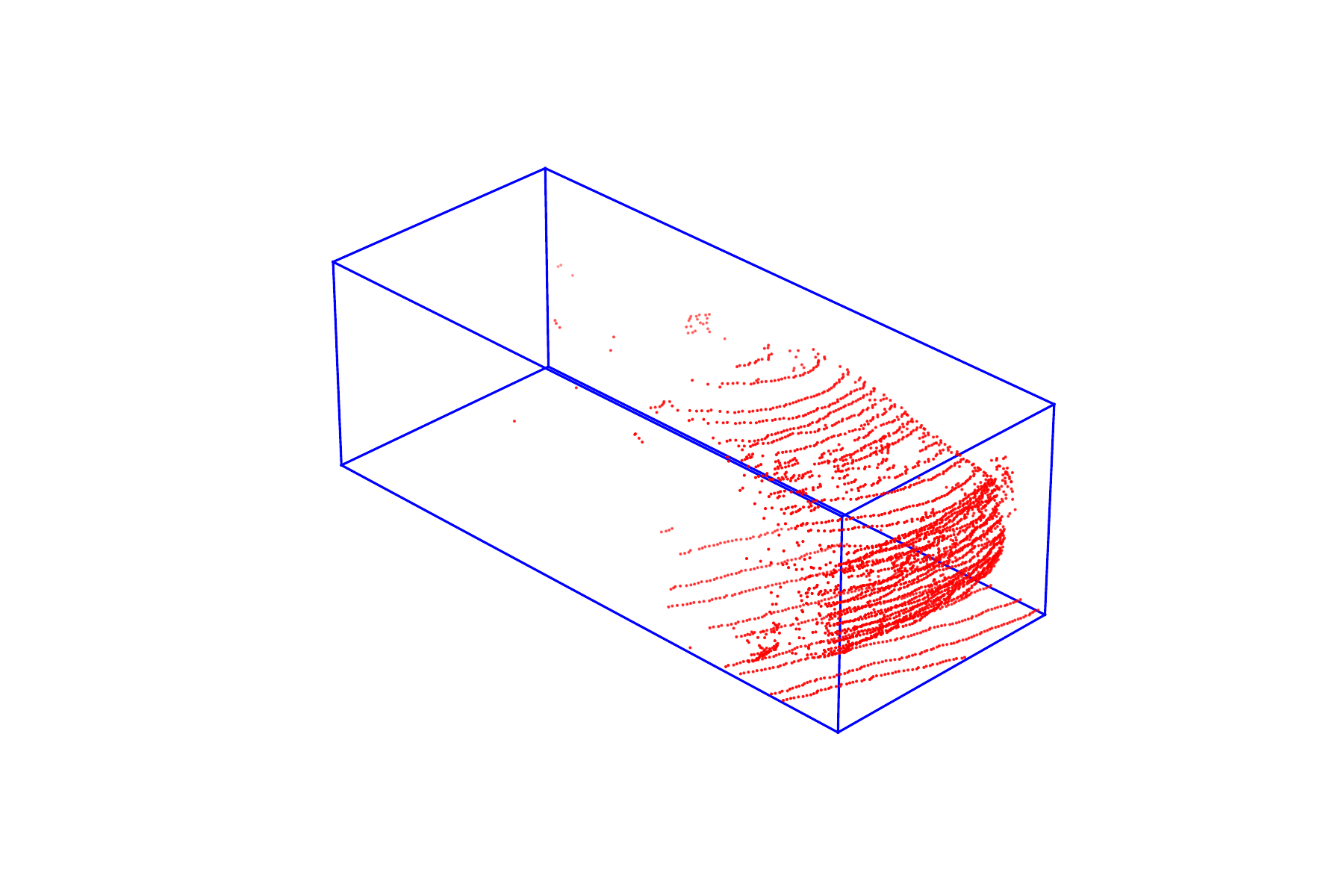}
        \caption{Car point cloud - solid-state LiDAR}
    \end{subfigure}
    \hfill
    \begin{subfigure}[b]{0.3\textwidth}
        \centering
        \includegraphics[width=\textwidth]{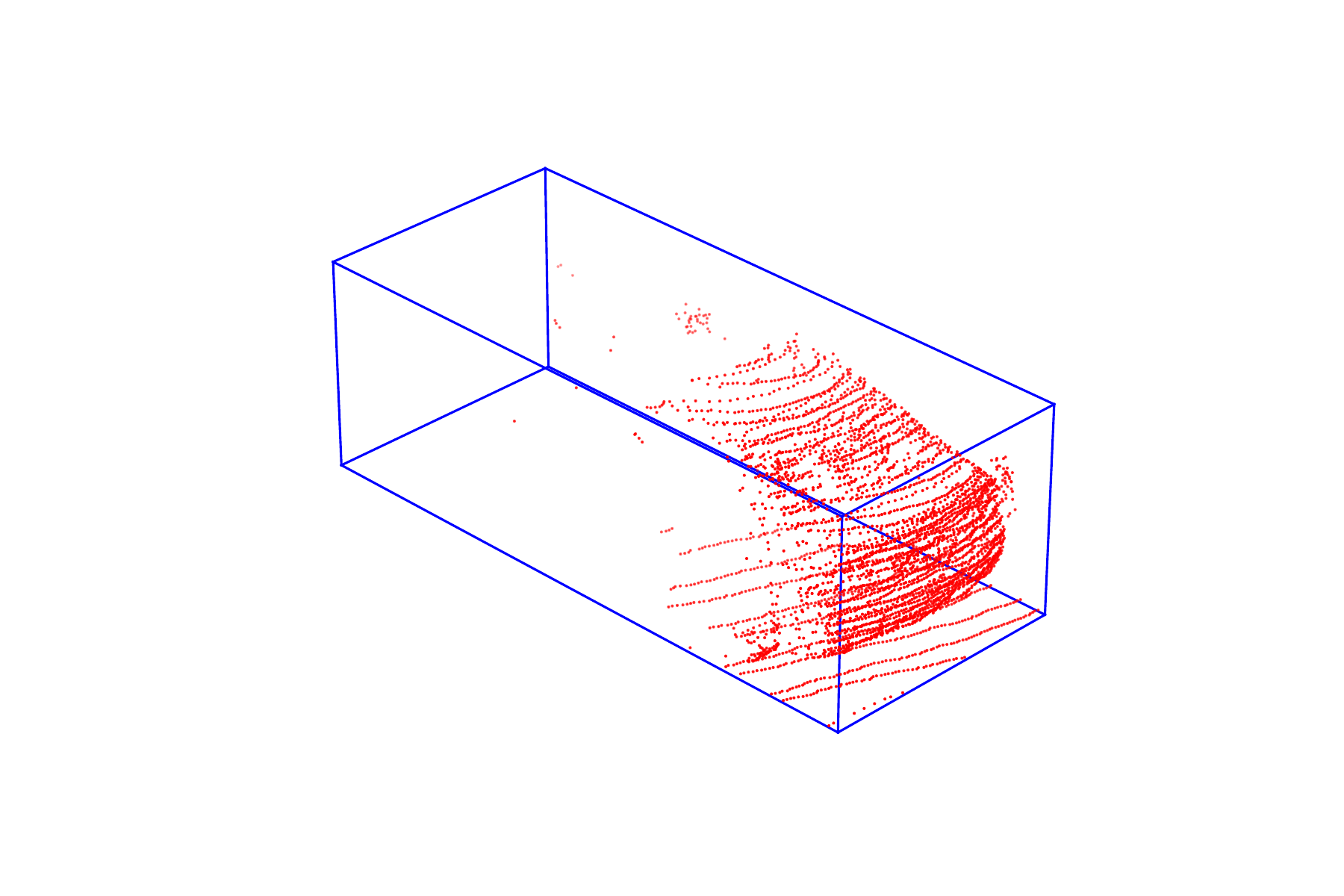}
        \caption{Car point cloud - merge of both LiDARs}
    \end{subfigure}
    
    \vskip\baselineskip  

    \begin{subfigure}[b]{0.3\textwidth}
        \centering
        \includegraphics[width=\textwidth]{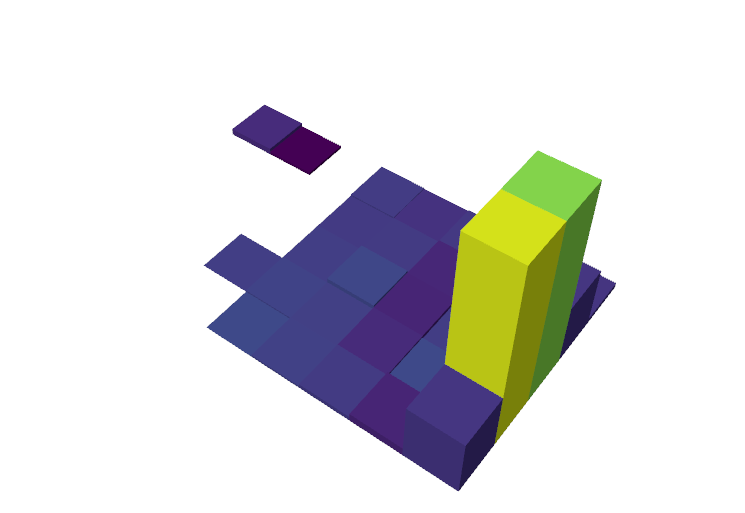}
        \caption{Car pillar representation - mechanical spinning LiDAR}
    \end{subfigure}
    \hfill
    \begin{subfigure}[b]{0.3\textwidth}
        \centering
        \includegraphics[width=\textwidth]{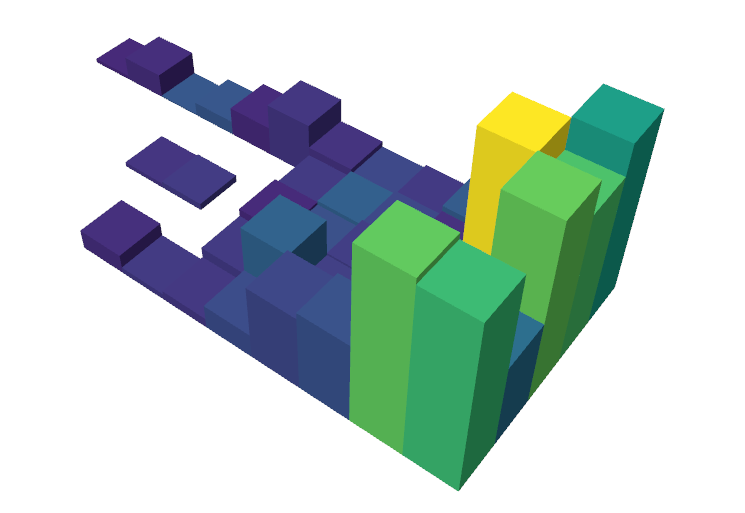}
        \caption{Car pillar representation - solid-state LiDAR}
    \end{subfigure}
    \hfill
    \begin{subfigure}[b]{0.3\textwidth}
        \centering
        \includegraphics[width=\textwidth]{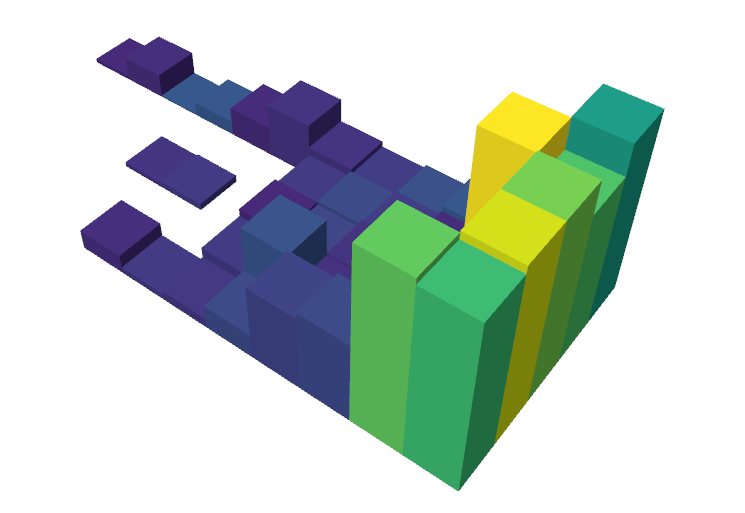}
        \caption{Car pillar representation - merge of both LiDARs}
    \end{subfigure}

    \caption{Car captured by LiDARs with different scanning mechanism (height of pillar - max intensity, color is max value in embedding)}
    \label{fig:points-voxels-visualization}
\end{figure*}

Subsequently, the disparities in LiDAR are reflected in the point cloud and its representations utilized in models. As shown in Figure \ref{fig:points-voxels-visualization} the same object (car) is represented differently in the point clouds by LiDARs with different scanning mechanisms. The visible difference is apparent to the unassisted eye. It is also apparent from the pillar representation that current state-of-the-art detection models were not designed for such cases. The question arises as to why the same physical object appears different for the model when recorded by different LiDARs. It is clear that the point cloud feature extraction layer must be enhanced to address the domain gap created by the growing diversity of the LiDAR sensors market. Until then, the intersensor LiDAR domain adaptation will remain a significant topic.

\subsubsection{Laser Source and Wavelength}
The majority of the popular sensors available on the market use pulsed-laser diodes as a laser source to emit laser beams at near-infrared wavelengths (typically 850-950 nm). The sole exception mentioned in our paper is the Hesai PandarGT, a solid-state LiDAR that utilizes short-wave infrared (1550 nm) beams emitted by a fiber laser. The 1550-nm wavelength exhibits a higher power limit than 905 nm for Class 1 eye safety, thereby enabling the laser to scan at greater distances. However, they are more susceptible to absorption and scattering in rain and fog compared to 905-nm systems. 

\subsection{Weather Conditions}
Adverse weather conditions, such as rain, fog, or snow, significantly degrade the quality of LiDAR data, as shown in \cite{heinzler2022LiDAR} by introducing noise and reducing the reflectivity of the points. Studies indicate a significant drop in model quality when trained on sunny-day data and tested in rainy conditions. The Waymo and nuScenes datasets, collected in varying weather conditions, show that models trained in clear weather perform poorly in rain or fog. The SPG paper \cite{xu2021spg} on domain adaptation highlights this as a significant adaptation challenge.

\subsection{Cultural and Geographic Factors}
Culture and geography have a significant impact on the types of vehicles in use around the world, directly affecting their size, design, and overall appearance. For instance, North American roads tend to feature larger SUVs, pickup trucks, and full-size sedans, while European cities are often dominated by smaller, more compact cars. In some parts of Asia, the prevalence of small sedans, scooters, rickshaws or motorcycles also significantly changes the overall makeup of road traffic.
The mean values of car bounding boxes in leading databases are noticeably different as can be seen in Tab.\ref{tab:car_dims}.
\begin{table}
    \centering
    \tabcolsep=0.11cm
    \begin{tabular}{cccc}
        Dataset  & Length [m] & Width [m] & Height [m] \\
        \hline
        KITTI    & 4.4    & 1.79  & 1.49   \\
        NuScenes & 4.61   & 1.95  & 1.73   \\
        Waymo    & 5.15   & 1.93  & 1.71   \\
    \end{tabular}
    \caption{Average car bounding box size in AV databases.}
    \label{tab:car_dims}
\end{table}
These region-specific preferences are further reinforced by local regulations (e.g., emissions laws, engine size taxes) and infrastructure constraints (e.g., narrow European streets vs. expansive American road networks), creating systematic shifts in the distribution of object sizes and shapes observed by LiDAR sensors. These variations introduce domain gaps (see Fig. \ref{fig:vehicle_size_datasets}) that can impede model generalization; a detector primarily trained in one region (where certain vehicle types are overrepresented) may struggle when deployed in areas with different vehicle shapes or sizes.

\subsection{Object labeling policy}
Another factor that could hinder domain adaptation results is the way datasets label objects. The most notable difference in labeling policies is the inclusion of vehicles' side view mirrors. Waymo dataset specifies that vehicle bounding boxes shall include side view mirrors, while NuScenes labeling specification excludes them. Although differences might seem minute, such metrics as IoU are very sensitive to small shifts and size differences. Such differences in data labeling policy results in errors not caused by errors in detections, but contradictory labeling guidelines. Figure \ref{fig:vehicle_size_datasets} demonstrates the extent of the vehicle sizes gap among datasets.

\begin{figure}[!h]
    \centering
    \includegraphics[width=\linewidth]{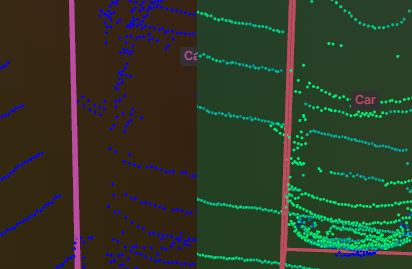}
    \caption{Inconsistencies in the inclusion of side view mirrors in the annotation could have a detrimental effect to the efficiency of domain adaptation between datasets. Waymo dataset (left) includes mirrors, while KITTI (right) doesn't.}
    \label{fig:side_view_mirrors}
\end{figure}

Furthermore, not all datasets have a labeling specification. KITTI dataset, in which side view mirrors are not included in the box, does not provide a labeling specification similar to the aforementioned Waymo and Nuscenes sets. Zenseact dataset also doesn't publish a detailed labeling specification, and more problematically, is not coherent in the inclusion of side view mirrors in annotations, meaning that some cars include them while other don't. The difference in side view mirror inclusion is illustrated by figure \ref{fig:side_view_mirrors}.

Other minor differences in labeling include crane arms, which are omitted in the NuScenes dataset if their height is above 1.5m, while Waymo labels such objects with a separate bounding box from the rest of the vehicle. Another difference lies in the case of humanoid mannequins: NuScenes requires them do be labeled as adult pedestrians, while Waymo doesn't count them as objects of interest. KITTI dataset, recorded in a city with a tram network, labels trams, while Waymo instructions exclude labeling of any rail vehicles.

Finally, differences in labeling policies are visible in the number and types of object classes. Waymo categorizes all motor vehicles with "vehicle" label, while KITTI differentiates between a "car", "truck" and "van". NuScenes and Zenseact datasets use multi-level categories, so that the user can select the granulation level of labels.

\begin{figure}[phtb]
    \centering
    \includegraphics[width=\linewidth]{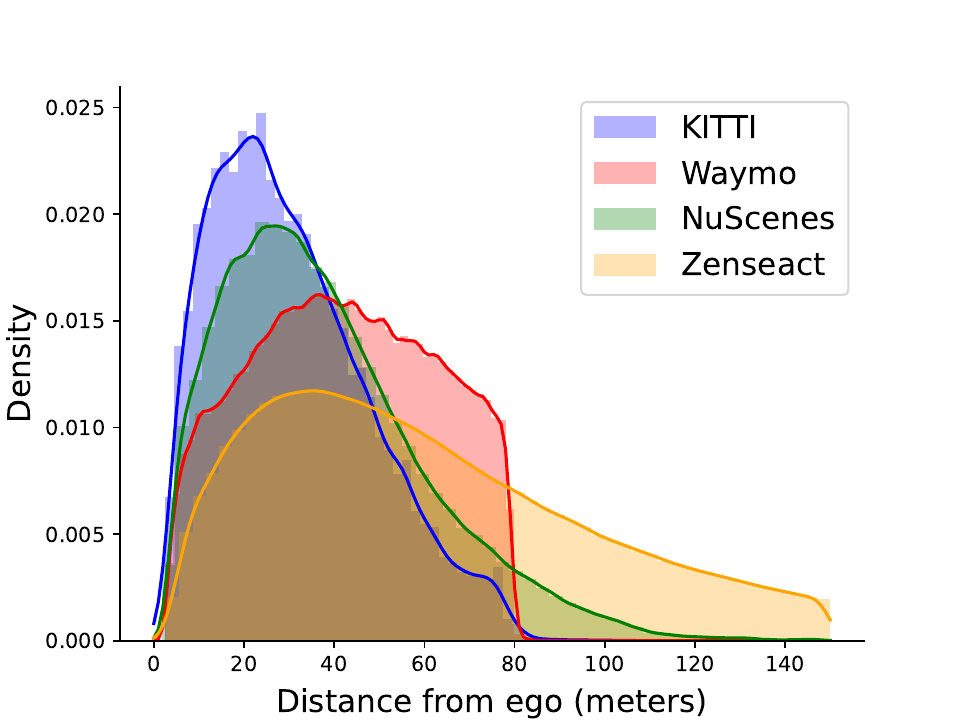}\\
    \caption{Distribution of distances of annotations from the ego vehicle. These distributions result both from the range of point clouds as well as from the labeling guideline.}
    \label{fig:dist_from_ego_datasets}
\end{figure}

Labeling guidelines specify the maximum distance of annotated objects from the ego vehicle. Datasets like NuScenes, KITTI and Zenseact require all objects visible in the image to be annotated, with Waymo being a notable exception where the area of interest is restricted to 80 meters from the ego vehicle, due to very low point cloud visibility above that mark. Figure \ref{fig:dist_from_ego_datasets} illustrates the differences in distributions of annotation distances.

\subsection{Additional considerations}
Cross-regional adaptations, such as the transfer of models from European datasets (KITTI) to US datasets (Waymo) \cite{train_in_germany}, reveal a significant decrease in performance due to differences in object sizes, frequently leading to predictions that do not meet the Intersection over Union (IoU) threshold. In these cases, implicit methods like Statistical Normalization (SN) prove effective by aligning feature distributions between domains to improve detection consistency. 

In addition, regional differences often introduce new challenges in the form of unique animals or obstacles that are specific to a given environment. For example, transitioning from European to US datasets may require adaptation to detect region-specific wildlife (e.g., deer or wild boar) or infrastructure variations. The detection and classification of such region-specific objects often require additional model adjustments or fine-tuning, which can further complicate cross-domain adaptation processes.

Furthermore, big differences in point cloud ranges (PCR) or detection region of interest (DRoI) between datasets contribute to loss of detection quality when training models on the merge of the datasets as shown in \cite{Zhang_2023_CVPR}, necessitating adjustments of the point cloud range and detection region of interest to maintain consistent object detection quality across domains.

\subsection{Existing datasets}
Many LiDAR datasets have been developed to support autonomous driving research, each offering unique sensor configurations and environmental contexts. Table \ref{tab:datasets} provides a comparative overview of the most popular datasets, focusing on sensor specifications and weather conditions important to domain adaptation strategies.

Each dataset varies in terms of sensor models used that differ in number of beams, fields of view (FoV), angular resolutions, capturing frequency, etc. Those differences result in different representations of the surrounding world. As the authors of \cite{fang2024LiDAR} show, current models struggle to generalize to the point clouds produced by LiDARs of different specifications.

Thus, one should analyze the target dataset from this point of view, and find the best matching source dataset for better domain adaptation performance.

\textbf{KITTI} \cite{kitti_dataset}.
The KITTI dataset, one of the first for autonomous driving research, uses the spinning Velodyne HDL-64E sensor with 64 beams, a vertical resolution of 0.4°, and a horizontal resolution of 0.08°. It contains data of driving environments in and around the city of Karlsruhe in Germany in clear weather. However, only the objects visible from the front-facing camera, featuring 90° horizontal field of view, were annotated.

\textbf{nuScenes} \cite{nuscenes_dataset}.
The nuScenes dataset uses a Velodyne HDL-32E sensor with 32 beams, vertical resolution of 1.33° and horizontal resolution of 0.32° rotating at 20Hz frequency. The lower resolution and number of beams compared to KITTI results in much sparser point clouds, which makes the task of domain adaptation in the case of nuScenes even more challenging. It includes 360° annotated data in urban scenes of Boston and Singapore in various weather conditions.

\textbf{Waymo} \cite{waymo_dataset}.
The Waymo Open dataset employs a high-resolution setup with multiple Velodyne LiDARs, including a top-mounted LiDAR with 64 beams, and 4 side-mounted short-distance LiDARs. This configuration allowed for dense environmental mapping across urban and suburban areas of the US. Unfortunately, the authors do not provide any details on the LiDAR models they used.

\textbf{Lyft Level 5} \cite{lyft_dataset}.
The Lyft Level 5 dataset includes data from multiple sensors, but specific LiDAR details, such as the exact sensor model and their resolutions, are not publicly available. The dataset covers diverse urban Palo Alto scenes with 360-degree annotations in clear weather conditions. Although 3 LiDARs were used, the point clouds in the dataset seem to be fairly sparse with $\approx$60K points per frame.

\textbf{Argoverse 1 \& 2} \cite{argoverse_dataset}. Another approach was used in the Argoverse dataset. The vehicle was equipped with two roof-mounted, rotating 32 beam LiDAR sensors. One on top of the other, each with 40° vertical field of view resulting in 30° vertical FoV overlap and a total of 50° vertical field of view. LiDAR range is up to 200 meters, roughly twice the range as the sensors used in nuScenes and KITTI. The authors state that on average, their LiDAR sensors produce a point cloud in each sweep with a density three times the density of the LiDAR sweeps in the nuScenes \cite{nuscenes_dataset} dataset, which is known to be quite sparse. The two LiDAR sensors rotate at 10 Hz and are out of phase, i.e. rotating in the same direction and speed but with an offset to avoid interference.

\textbf{Zenseact Open Dataset} \cite{zenseact_dataset}. This dataset features aggregated point clouds from 3 Velodyne LiDAR sensors, the top one with 128 beams, and two side ones capturing 16 beams. The dataset is the densest among the ones mentioned in this paper, providing  $\approx$254K points per frame. Similarly to KITTI, the Zenseact dataset contains only annotations of the objects visible from the front-facing camera that has a horizontal field of view of 120°.

\textbf{PandaSet} \cite{xiao2021pandaset}. PandaSet utilizes 2 Hesai LiDARs of different nature. The Pandar64 sensor on the top offers 64 beams with a vertical angular resolution ranging from 0.17° to 6° and a horizontal angular resolution of 0.2°. The second PandarGT sensor is so-called solid-state LiDAR. It uses a solid-state laser and a photodetector to measure distances. Unlike mechanical LiDAR, solid-state LiDAR does not have moving parts, making it more reliable and less susceptible to failure. The dataset provides high-density point clouds captured in complex urban environments in San Francisco and the El Camino Real route.

\textbf{ONCE} \cite{mao2021one}. The dataset features a million scenarios from multiple Chinese cities captured by a 40-beam top LiDAR with 40° vertical FoV and 360° horizontal FoV. It includes data registered in clear, cloudy, and rainy weather. The average of $\approx$70K points per frame makes the ONCE point clouds sparser than most of the previously mentioned datasets.

\textbf{AiMotive} \cite{Matuszka2022aiMotiveDA}. This dataset contains scenes from USA, Austria, and Hungary recorded by a 64-beam LiDAR. The distinguishing feature of the aiMotive dataset is long-range 3D annotations reaching up to 200 meters away from the ego vehicle, allowing for long-range perception. Unfortunately, the information on the LiDAR used in the dataset is not publicly available.

\textbf{ApolloScape} \cite{Wang2018TheAO}. The ApolloScape dataset adopts the Riegl VMX-1HA acquisition system. The system features two VUX-1HA 360° horizontal FoV laser scanners. Although the specific sensor characteristics are not mentioned, the authors state that their system is able to achieve higher density point clouds than the ones captured by the common Velodyne HDL-64E LiDAR. The dataset has many complex scenes recorded in Beijing, China, perfectly suited to the dense urban environment of eastern Asian countries.

\textbf{A2D2} \cite{geyer2020a2d2}. A2D2 is another dataset for autonomous vehicle applications recorded in Germany. It utilizes five synchronized Velodyne VLP-16 LiDARs having 16 beams and capturing data at 10Hz frequency, which seems enough to create colored 3D maps. Although being good at short-range scanning, the setup struggles to capture long-range objects. The dataset provides 3D bounding boxes for the points within the field of view of the front-center camera having 60° horizontal FoV.

\textbf{ParisLuco3D} \cite{10508037}. ParisLuco3D is a dataset captured around the Luzembourg Garden in Paris. The LiDAR data was recorded by the same sensor as the nuScenes dataset, with the only difference being the capturing frequency. While the nuScenes sensor rotated at 20Hz, the ParisLuco3D setup used the LiDAR at a 10Hz rotating speed, which allowed for the use of a lower HAR, resulting in a twofold increase in the density of the point cloud per frame. This is beneficial for the object detection task.

\textbf{LIBRE} \cite{9304681}. LIBRE is a dataset recorded by 10 different LiDARs from a vehicle driving on public urban roads around Nagoya University, Japan. This includes Velodyne LiDARs: VLS-128, HDL-64S2, HDL-32E, VLP-32C, and VLP-16; Pandar LiDARs: Pandar64, Pandar40P; Ouster LiDARs: OS1-64, OS1-16; RoboSense LiDAR: RS-Lidar32. More details on sensor configurations can be found in \cite{9304681}. The dataset features scenarios of a trajectory recorded by each sensor at three time periods during the day.

\textbf{A*3D} \cite{9197385}. The A*3D dataset is a diverse dataset that covers the whole variety of the Singapore road environment (unlike nuScenes, which contains only a small portion of it). It features 3D annotations of the objects visible from any of the frontal-view cameras like KITTI, Zenseact and A2D2. 

Each of these datasets brings unique LiDAR configurations and environmental diversity, significantly impacting domain adaptation strategies. Differences in beam count, vertical and horizontal angular resolutions, and field of view influence the adaptability of models trained on a single dataset to new domains.

\begin{table*}[htbp]
    \centering
    \tiny
    \renewcommand{\arraystretch}{1.1} 
    \begin{tabularx}{\textwidth}{|l|X|X|p{0.12\linewidth}|X|X|X|X|X|c|}
        \toprule
        \textbf{Dataset} & \textbf{Country} & \textbf{Env} & \textbf{Sensors} & \textbf{SM} & \textbf{VAR} & \textbf{HAR} & \textbf{Beams} & \textbf{WC} & \textbf{MPF} \\
        \midrule
        KITTI       & Germany & Urban, suburban, highways & 1xVelodyne HDL-64E & 1xMS & 0.4 & 0.08 & 1x64 & Clear & 120K \\
        \hline
        nuScenes    & USA, Singapore & Urban & 1xVelodyne HDL-32E & 1xMS & 1.33 & 0.32 & 1x32 & Clear, cloud, rain & 34K \\
        \hline
        Waymo       & USA & Urban, suburban & 1x mid-range LiDAR \newline 4x short-range LiDARs & 5xMS & 0.31 \newline - & 0.16 \newline -  & 1x64 \newline - & Clear, cloud, rain & 177K \\
        \hline
        Lyft Level 5 & USA & Urban & 1xTop LiDAR* \newline 2xFront LiDAR* & 3xMS & - & - & 1x64 \newline 2x40  & Clear & 62K \\
        \hline
        Argoverse 1 & USA & Urban, suburban & 2xVelodyne VLP-32C & 2xMS & 0.33\textsuperscript{!} & 0.2 & 2x32 & Diverse* & 107K \\
        \hline
        Argoverse 2 & USA & Urban, suburban & 2xVelodyne VLP-32C & 2xMS & 0.33\textsuperscript{!} & 0.2 & 2x32 & Diverse* & 107K \\
        \hline
        Zenseact    & 14 European countries & Urban, suburban, highways & 1xVelodyne VLS-128 \newline 2xVelodyne VLP-16 & 3xMS & 0.11\textsuperscript{!} \newline 2 & 0.2 \newline 0.2 & 1x128 \newline 2x16 & Clear, cloud, rain, snow  & 254K \\
        \hline
        PandaSet    & USA & Urban, suburban & 1xPandar64 \newline 1xPandarGT & 1xMS \newline 1xSS & 0.17\textsuperscript{!} \newline 0.07\textsuperscript{!} & 0.2 \newline 0.1 & 1x64 \newline 1x150 & Clear & 169K \\
        \hline
        ONCE        & China & Urban & 1xTop LiDAR* & 1xMS & 0.33\textsuperscript{!} & 0.2 & 1x40 & Clear, cloud, rain & 70K \\
        \hline
        aiMotive    & USA, Austria, Hungary & Urban & 1xTop LiDAR* & 1xMS & - & - & 1x64 & Clear, cloud, rain, glare & 120K \\
        \hline
        ApolloScape & China & Urban & 2xRIEGL VUX-1HA & 2xMS & - & - & - & Diverse* & - \\
        \hline
        A2D2        & Germany & Urban, suburban, highways & 5xVelodyne VLP-16 & 5xMS & 2 & 0.2 & 5x16 & Clear, cloud, rain & - \\
        \hline
        ParisLuco3D & France & Urban & 1xVelodyne HDL-32E & 1xMS & 1.33 & 0.16 & 1x32 & - & 68K \\
        \hline
        LIBRE       & Japan & Urban & 10 sensors from 4 different producers & 10xMS & Var & Var & Var & Clear, cloud & Var \\
        \hline
        A*3D        & Singapore & Urban, suburban & 1xVelodyne HDL-64ES3 & 1xMS & 0.33-0.5 & 0.17 & 1x64 & Clear, cloud, rain & 133K \\
        \bottomrule
    \end{tabularx}
    \caption{
        Each dataset varies in terms of recordings location, sensor models of different scanning mechanisms (SM) used (MS: mechanical spinning, SS: solid-state), vertical and horizontal angular resolutions (VAR and HAR), number of beams, weather conditions (WC), and mean points per frame (MPF). These distinctions play a crucial role in how well models trained on one dataset can adapt to another.
        \newline (-) - no information is provided.
        \newline (*) - details are not specified
        \newline (\textsuperscript{!}) - minimum (finest) resolution, a sensor has variable angle delta between beams
        \newline (Var) - variable, the scenes were recorded by different setups
    }
    \label{tab:datasets}
\end{table*}

\section{Effective domain transfer}
When creating a new labeled LiDAR dataset, an effective strategy is to begin with a model already trained on data that closely matches the sensor setup and environmental conditions of the target domain. For instance, if you plan to capture Polish road data using a Velodyne VLS128 sensor at 10Hz, it’s more straightforward to adapt a model trained on the Zenseact dataset—collected with the same or very similar LiDAR settings—than to start with datasets like nuScenes or Waymo, which use different sensors in different environments.

\subsection{Domain Transfer Across Regions} 
In this subsection we consider domain transfer across the regions of the greatest autonomous vehicles development potential.

\textbf{Europe - USA}: This transfer requires careful adjustment due to significant differences in road infrastructure and driving habits. For example, European cities often have narrower roads, more pedestrian zones, and different traffic patterns compared to the USA’s grid-like road systems and wider lanes.

\textbf{USA - Asia}: Domain transfer to regions like Japan or South Korea introduces even greater complexity due to highly urbanized environments, dense traffic, and specific road regulations unique to Asian countries. In addition, differences in signage and lane usage add to adaptation challenges.

\textbf{Europe - Asia}: Similarly to the USA-to-Asia case, transferring models from European roads to Asian domains involves overcoming variations in infrastructure, signage, and traffic patterns. However, urbanization levels and driving behaviors in certain European countries (e.g., densely populated areas like Germany or the Netherlands) may be more comparable to Asian cities.

\subsection{Intra-Regional Domain Transfer}
Domain transfer within a given region is often less complex when adapting LiDAR-based autonomous vehicle models, thanks to shared road systems, regulations, and vehicle types. For example, within Europe, a LiDAR model trained on data from Germany can be adapted for operation in Poland, France, or Spain without substantial modifications, as these countries share many driving patterns. A similar scenario is evident in the United States, where a LiDAR model developed in one state can often be transferred to another with only minor adjustments to account for local variations in road infrastructure or climate conditions. In both cases, sensor setups generally remain consistent and modest adaptations, such as fine-tuning environmental models to address climate differences or local topographical features, are typically sufficient to ensure reliable performance.

\subsection{Conclusion}
Domain transfer is easiest within regions with similar driving environments and infrastructure, such as states in the United States or countries within Europe. Transfers between continents, such as from the United States to Europe or Asia, are more challenging due to infrastructure differences, environmental variations, and regulatory diversity. For the most effective results, models should be trained and fine-tuned on datasets that closely match the target domain in terms of sensor configuration and environmental conditions.

\section{Additional experiments}

{\bf \noindent Size of the target domain.}
In this experiment, we trained the model from scratch using prepared small subsets of the KITTI, NuScenes and Waymo datasets. The detector performance we obtained is shown in Tab. \ref{tab:result_small_ds_scratch}. Prior to these experiments, we increased the number of epochs in the training setup from 40 to 4,000 so that the model weights could establish themselves. This was due to the fact that when using a small amount of data during one epoch, the model weights update more slowly than when there is more data. Therefore, with a small portion of data, a training consisting of more epochs is needed.

The results obtained in this way are unexpectedly high, but markedly lower than when learning on the entire training set. There are considerable discrepancies in the AP metric value calculated with an IoU threshold of 0.5 and a threshold of 0.7, indicating that the detector is capable of locating the object with reasonable precision, but is unable to accurately estimate its size and precise location. It is worth noting that these results are lower than those achieved when the model was initially trained on a different domain. This suggests that when a company has limited access to annotated data in the development period, it is more effective to first train the model on the publicly available datasets for commercial purposes and then apply post-training using the available samples from the currently possessed data.


\begin{table*}[!htbp]
\caption{Comparison of model results with post-training using only 10 frames from the target dataset. The evaluation is shown for both source (upper row) and target (lower row) datasets. Result before / sign is without post training and after / sign is after post training.}
\centering
\label{tab:result_small_ds_scratch}
\scriptsize
\begin{tabular}{@{}ccccccc@{}}
\toprule
\multirow{2}{*}{\diagbox{\textbf{S.}}{\textbf{T.}}} 
  & \multicolumn{2}{c}{\textbf{KITTI}} 
  & \multicolumn{2}{c}{\textbf{NuScenes}} 
  & \multicolumn{2}{c}{\textbf{Waymo}} \\
\cmidrule(lr){2-3} \cmidrule(lr){4-5} \cmidrule(lr){6-7}
  & $\bm{AP_{0.5}}$ & $\bm{AP_{0.7}}$ & $\bm{AP_{0.5}}$ & $\bm{AP_{0.7}}$ & $\bm{AP_{0.5}}$ & $\bm{AP_{0.7}}$ \\
\midrule

\textbf{K.} 
  & -- & -- & 96.8 / 86.8 & 88.0 / 17.7 & 96.8 / 94.0 & 88.0 / 51.8 \\
  &    &    & 23.9 / 28.3 & 7.3 / 14.2  & 52.9 / 66.8 & 11.2 / 46.0 \\
\midrule

\textbf{N.} 
  & 42.3 / 25.9 & 26.0 / 5.1 & -- & -- & 42.3 / 24.9 & 26.0 / 11.1 \\
  & 80.6 / 93.5 & 23.6 / 79.9 &    &    & 44.5 / 65.6 & 18.6 / 49.4 \\
\midrule

\textbf{W.} 
  & 78.7 / 57.3 & 67.2 / 17.6 & 78.7 / 13.8 & 67.2 / 5.7 & -- & -- \\
  & 88.9 / 96.7 & 7.7 / 83.0  & 19.4 / 30.9 & 5.4 / 12.3 &    &    \\
\bottomrule
\end{tabular}
\end{table*}

{\bf \noindent Post-training on single vs. multiple domains.} Additionally, we evaluated the model’s performance on the source dataset after post-training on the target domain. The primary objective was to determine whether domain adaptation affects the model’s ability to perform on the original domain. The results, presented in Tab.\ref{tab:post_KNW_source_target}, indicate a noticeable decline in performance on the source dataset—showing a slight drop in $AP_{0.5}$, and a more significant decrease is observed for $AP_{0.7}$. This suggests that while the model can still detect objects from the source domain, its precision is substantially reduced. These findings highlight the model’s susceptibility to catastrophic forgetting, where adapting to a new domain leads to a loss of previously learned knowledge. To mitigate this issue, more advanced techniques from continual learning could be applied to preserve the model’s original performance after domain adaptation.

{\bf \noindent Post-training strategy.} To further investigate the impact of different post-training strategies in the context of domain adaptation, we conducted an additional experiment focused specifically on the pedestrian class again using only 10 frames. This was motivated by the typically observed variability in performance across semantic classes in 3D object detection problem. Pedestrians often represents a challenging category due to their diverse appearances and frequent occlusions.

The experiment evaluates the effectiveness of various post-training (fine-tuning) strategies in enhancing cross-domain generalization for the pedestrian class. The corresponding results are summarized in Table \ref{tab:pedestrian}. Interestingly, for the pedestrian category, the model achieves even higher performance on the KITTI dataset after post-training than it does on the target domain itself. This counterintuitive result may be attributed to the greater diversity of pedestrian instances in the Waymo and nuScenes datasets, which likely contributes to a more robust and generalized pedestrian representation during adaptation. In contrast, for the vehicle class, the post-training domain adaptation results in a marginal drop in performance compared to the target domain, suggesting that the benefits of post-training may be class-dependent and influenced by the variability present in the source and target datasets.

\begin{table}[h]
 \caption{Performance of pedestrian domain adaptation using only 10 frames on PV-RCNN with $AP_{0.5}$ metric.}
  \scriptsize
  \centering
  \label{tab:pedestrian}
  \tabcolsep=0.11cm
  \begin{tabular}{|c|c|c|}
    \hline
    \textbf{Adaptation task}               & \textbf{Method}   & $\bm{AP_{0.5}}$                        \\
       \hline\hline
        \multirow{5}{*}{\textbf{NuScenes $\rightarrow$ KITTI}}  & Target domain                       & 34.44                  \\
                                                                & Source only                         & 2.88                 \\                   
                                                                & 10 f. + const LR (ours)       &  43.25           \\
                                                                & 10 f. + LR fading (ours)            & 41.76      \\
                                                                & 10 f. + L2-SP (ours)                & 44.07      \\
        \hline
        \multirow{5}{*}{\textbf{Waymo $\rightarrow$ KITTI}}    & Target domain                       & 34.44                \\
                                                                & Source only                         & 46.66             \\
                                                                &  10 f. + const LR (ours)       &  51.38          \\
                                                                & 10 f. + LR fading (ours)            & 50.92      \\
                                                                & 10 f. + L2-SP (ours)                & 52.83     \\
\hline        \multirow{5}{*}{\textbf{Waymo $\rightarrow$ NuScenes}}    & Target domain                       & 22.48            \\
                                                                & Source only                         & 6.32              \\
                                                                & 10 f. + const LR (ours)              &  12.88     \\
                                                                & 10 f. + LR fading (ours)            & 14.36        \\
                                                                & 10 f. + L2-SP (ours)                &  13.23       \\
        \hline
  \end{tabular}
\end{table}

\begin{table*}[!htb]
  \caption{Diverse vs random fine-tuning frame selection. TED detector, $AP_{0.7}$ metric. The results for the random selection correspond to three independent runs, with the values in parentheses indicating the average performance across these three experiments.}
  \centering
  \label{tab:10_100_post-training_ap70_random}
  \tabcolsep=0.15cm
  \footnotesize
  \begin{tabular}{|c|c|c|c|c|}
    \hline
    \diagbox{\textbf{S.}}{\textbf{T.}} & \textbf{Frames} & \textbf{KITTI} & \textbf{NuScenes} & \textbf{Waymo} \\
    \hline\hline

    \multirow{4}{*}{\textbf{KITTI}} 
        & +10 (div.)  & --   & 14.4 & 51.6 \\
        & +100 (div.) & --   & 17.4 & 58.2 \\
        & +10 (ran.)  & --   & 12.9 / 13.2 / 13.5 (13.2) & 49.0 / 45.9 / 49.6 (48.2) \\
        & +100 (ran.) & --   & 17.7 / 17.2 / 18.8 (17.9) & 58.0 / 56.2 / 57.1 (57.1) \\

    \hline
    \multirow{4}{*}{\textbf{NuScenes}} 
        & +10 (div.)  & 81.1 & -- & 49.0 \\
        & +100 (div.) & 82.5 & -- & 57.6 \\
        & +10 (ran.)  & 78.6 / 81.5 / 77.6 (79.2) & -- & 48.4 / 46.4 / 48.0 (47.6) \\
        & +100 (ran.) & 83.5 / 83.7 / 82.3 (83.2) & -- & 57.9 / 57.2 / 57.8 (57.7) \\

    \hline
    \multirow{4}{*}{\textbf{Waymo}} 
        & +10 (div.)  & 82.0 & 11.0 & -- \\
        & +100 (div.) & 85.1 & 17.6 & -- \\
        & +10 (ran.)  & 78.5 / 83.3 / 80.9 (80.9) & 12.4 / 9.5 / 12.0 (11.3) & -- \\
        & +100 (ran.) & 85.4 / 85.6 / 85.2 (85.4) & 16.3 / 17.4 / 18.5 (17.4) & -- \\

    \hline
  \end{tabular}
\end{table*}

{\bf \noindent Impact of the samples selection.} 
In last experiment, we examined how the choice of sample selection strategy influences the effectiveness of domain adaptation. Specifically, we compared the performance of the model when adapted using randomly selected samples versus samples selected to maximize diversity. To evaluate the consistency of the random selection approach, we performed three independent runs using different random subsets. This allowed us to observe the variation in results and compare it with the performance achieved through diverse sampling. The detailed results of this comparison are provided in Tab. \ref{tab:10_100_post-training_ap70_random}.


As can be observed, when using 10 samples, the performance variation between the worst and best results reaches up to 5\%, indicating a relatively high sensitivity to sample selection. In contrast, with 100 samples, the variation is significantly reduced, suggesting greater stability and robustness in the adaptation process. This demonstrates that when a larger and more representative subset is used for domain adaptation, the specific choice of samples becomes less critical. In the context of autonomous driving, this implies that effective adaptation to a new target domain can be achieved by fine-tuning the model on a small set of carefully selected scenarios from a new ODD. These scenarios should include representative examples of all relevant classes and capture a range of distances from the ego vehicle to ensure comprehensive coverage of typical driving conditions.




\end{appendices}

\end{document}